\documentclass[10pt,twocolumn,letterpaper]{article}

\usepackage{iccv}
\usepackage{times}
\usepackage{epsfig}
\usepackage{graphicx}
\usepackage{subfigure}
\usepackage{amsmath}
\usepackage{amssymb}
\usepackage{booktabs}  % \toprule, \midrule, \bottomrule
\usepackage{bbding}    % \Checkmark, \XSolid, \CheckmarkBold, \XSolidBold, \XSolidBrush
\usepackage{multirow,enumitem}

\newcommand{\comment}[1]{}
\DeclareMathOperator*{\argmax}{arg\,max}

\usepackage{xcolor}
\usepackage[pagebackref=true,breaklinks=true,letterpaper=true,colorlinks,bookmarks=false]{hyperref}

\def\bfa#1{\textbf{\color{red}{#1}}}
\def\bfb#1{\textbf{\color{blue}{#1}}}

\makeatletter
\def\ps@myheadings{%
    \let\@oddfoot\@empty\let\@evenfoot\@empty
    \def\@evenhead{\thepage\hfil\slshape\leftmark}%
    \def\@oddhead{{\slshape\rightmark}\hfil\thepage}%
    \let\@mkboth\@gobbletwo
    \let\sectionmark\@gobble
    \let\subsectionmark\@gobble
    }
  \if@titlepage
  \renewcommand\maketitle{\begin{titlepage}%
  \let\footnotesize\small
  \let\footnoterule\relax
  \let \footnote \thanks
  \null\vfil
  \vskip 60\p@
  \begin{center}%
    {\LARGE \@title \par}%
    \vskip 3em%
    {\large
     \lineskip .75em%
      \begin{tabular}[t]{c}%
        \@author
      \end{tabular}\par}%
      \vskip 1.5em%
    {\large \@date \par}%       % Set date in \large size.
  \end{center}\par
  \@thanks
  \vfil\null
  \end{titlepage}%
  \setcounter{footnote}{0}%
}
\else
\renewcommand\maketitle{\par
  \begingroup
    \renewcommand\thefootnote{\@fnsymbol\c@footnote}%
    \def\@makefnmark{\rlap{\@textsuperscript{\normalfont\color{black}\@thefnmark}}}%
    \long\def\@makefntext##1{\parindent 1em\noindent
            \hb@xt@1.8em{%
                \hss\@textsuperscript{\normalfont\@thefnmark}}##1}%
    \if@twocolumn
      \ifnum \col@number=\@ne
        \@maketitle
      \else
        \twocolumn[\@maketitle]%
      \fi
    \else
      \newpage
      \global\@topnum\z@   % Prevents figures from going at top of page.
      \@maketitle
    \fi
    \thispagestyle{plain}\@thanks
  \endgroup
  \setcounter{footnote}{0}%
}
\makeatother

\makeatletter
\newcommand\fs@nobottomruled{\def\@fs@cfont{\bfseries}\let\@fs@capt\floatc@ruled
  \def\@fs@pre{}% \hrule height.8pt depth0pt \kern2pt
  \def\@fs@post{}% Formerly \def\@fs@post{\kern2pt\hrule\relax}%
  \def\@fs@mid{\kern2pt\hrule\kern2pt}%
  \let\@fs@iftopcapt\iftrue}
\makeatother

\iccvfinalcopy % *** Uncomment this line for the final submission

\def\iccvPaperID{5913} % *** Enter the ICCV Paper ID here

\ificcvfinal\fi

\begin{document}

%%%%%%%%% TITLE
\title{From Saliency to DINO: Saliency-guided Vision Transformer for Few-shot Keypoint Detection}
% \title{Detecting Few-shot Keypoints via Transformer Guided by Saliency Prior}
% \title{Detecting Few-shot Keypoints via Prior-mask Guided Vision Transformer}
% \title{Detecting Few-shot Keypoints by Saliency Prior Guided Vision Transformer}
% \title{From Saliency to Token Attentiveness: Detecting Few-shot Keypoints via Prior Map Guided Transformer}
% \title{From Saliency to DINO: Detecting few-shot keypoints with saliency priors}

\author{Changsheng Lu$^{1}$, Hao Zhu$^{1}$, Piotr Koniusz$^{2,1}$\\
$^{1}$The Australian National University \quad
 $^2$Data61/CSIRO\\
%College of Engineering and Computer Science, Australian National University\\
{\tt\small \{ChangshengLuu,allenhaozhu\}@gmail.com, Piotr.Koniusz@data61.csiro.au}
% \and
}

\maketitle
\ificcvfinal\thispagestyle{empty}\fi

%%%%%%%%% ABSTRACT
\begin{abstract}
  % Crucial problems:
  % 1) Most of existing FSL approaches use a simple CNN to extract deep embeddings, while neglects the foreground patch relations which could serve as a cue to help similarity learning. This stems from the local receptive field of convolutional kernel. Moreover, CNN is good at modeling local features while in short of capturing long-range relations.
  % 2) ViT features has such ability but receptive field is too wide. And irrelevant feature out of region of interest could potentially affects feature learning.
  % Three major things we have done in this work: i) Design a good SalViT which focuses relation learning on saliency region; ii) Augment CNN feature with SalViT feature, increasing the feature representation power for FSL; iii) FSKD and FSL.
  Unlike current deep keypoint detectors that are trained to recognize limited number of body parts, few-shot keypoint detection (FSKD) attempts to localize any keypoints, including novel or base keypoints, depending on the reference samples. FSKD requires the semantically meaningful relations for keypoint similarity learning to overcome the ubiquitous noise and ambiguous local patterns. One rescue comes with  vision transformer (ViT) as it captures long-range relations  well. However, ViT may model irrelevant features outside of the region of interest due to the global attention matrix, thus degrading similarity learning between support and query features. In this paper, we present a novel saliency-guided vision transformer, dubbed SalViT, for few-shot keypoint detection. Our SalViT enjoys a uniquely designed masked self-attention and a morphology learner, where the former introduces saliency map as a soft mask to constrain the self-attention on foregrounds, while the latter leverages the so-called power normalization to adjust morphology of saliency map, realizing ``dynamically changing receptive field''. Moreover, as salinecy detectors add computations, we show that attentive masks of DINO transformer can replace saliency. On top of SalViT, we also investigate i) transductive FSKD that enhances keypoint representations with unlabelled data and ii) FSKD under occlusions. We show that our model performs well on five public datasets and achieves $\sim\!10\%$ PCK higher than the normally trained model under severe occlusions.
\end{abstract}

%%%%%%%%% BODY TEXT
\section{Introduction}\label{sec:introduction}

\begin{figure}[!t]
\vspace{-0.3cm}
  \centering
  \includegraphics[width=\linewidth]{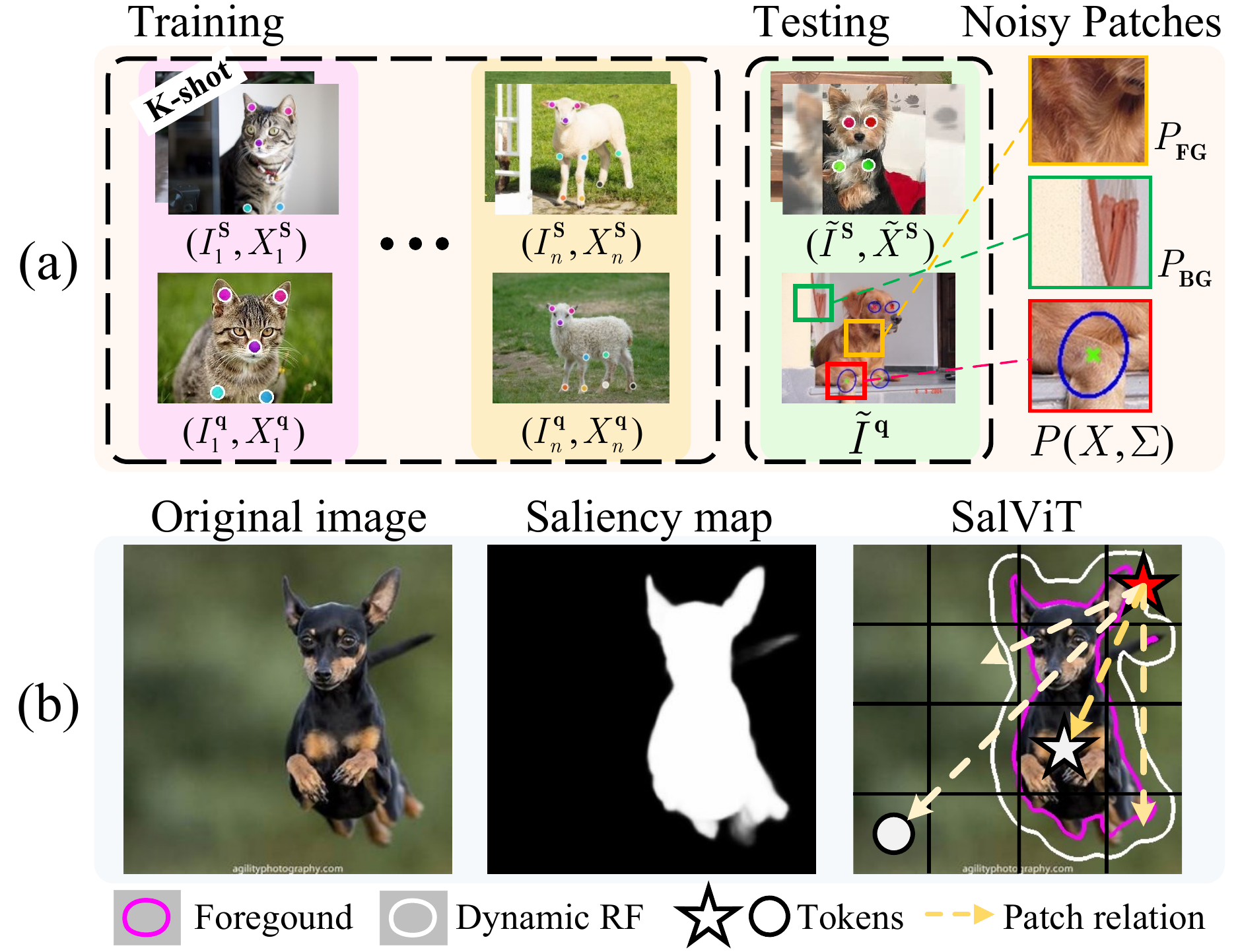}
  \caption{Few-shot keypoint detection (FSKD) and patch relations guided by saliency. (a) We train and test FSKD model via episodes, each containing a support image $I^{\text{s}}$, support keypoints $X^{\text{s}}$, and the query image $I^{\text{q}}$. The goal is to infer the keypoints $X$ in $I^{\text{q}}$ that correspond to $X^{\text{s}}$. $P_{\text{FG}}$ and $P_{\text{BG}}$ show the noisy foreground and background patches. $P(X, \Sigma)$ shows the predicted keypoint $X$ with uncertainty $\Sigma$ by our model (\emph{best viewed in zoom}); (b) The saliency map indicates the foreground patches. Our SalViT can enhance the relation learning for the pairwise salient patches (or ``tokens'') while suppressing non-salient pairs.
  % The saliency map is used to indicate the foreground patches in our SalViT. As a result, SalViT could enhance the relation learning for the pairwise salient patches (or ``tokens'') while weaken that of un-salient pairs. Such relation would help keypoint inference in FSKD.
  %and served as soft mask in our masked self-attention (M-SA). As a result, M-SA can enhance the relation learning for salient patches (or ``tokens'') while weaken that of unsalient ones, which would help to infer keypoint similarity in FSKD.
  }
  \label{fig:fskd-salvit}
  \vspace{-0.3cm}
\end{figure}

The deep fully-supervised learning relies on large amounts of labelled data for training and cannot generalize to novel classes well. In contrast, humans learn new concepts from few samples, which motivates the need for few-shot learning (FSL) \cite{wang2020generalizing} in various tasks such as image classification \cite{sung2018learning,gidaris2018dynamic,snell2017prototypical,vinyals2016matching,koch2015siamese}, object detection \cite{kang2019few,fan2020few,wang2020frustratingly,zhang2021hallucination}, segmentation \cite{li2020fss,li2021adaptive}, and keypoint detection \cite{Lu_2022_CVPR,ge2021metacloth}. The emerging few-shot keypoint detection (FSKD) localizes keypoints in query images given one or a few support images with keypoint annotations (Fig.~\ref{fig:fskd-salvit}(a)). The hardest FSKD setting (which we use among other settings)  is that both the base keypoints/species in training are \emph{disjoint} with the novel keypoints/species in testing, which is challenging due to the large domain shifts. %in keypoints and species.

%The hardest setting (which we use among other settings) due to the domain shift in keypoints and species of FSKD is that both the base keypoints and species in training are \emph{disjoint} with the novel ones in testing.
% which is quite challenging due to the large domain shift in keypoints and species (\eg seen \vs unseen). %Generally speaking, FSKD is a versatile keypoint meta-learner, which is expected to detect arbitrary keypoints as long as providing the support keypoints.

Compared to general FSL, FSKD shares two common challenges: 1) generalization to unseen classes; 2) low number of samples. Though seminal works~\cite{vinyals2016matching,snell2017prototypical,sung2018learning} address some of such issues, FSKD faces additional challenges. Firstly, keypoints represent semantic local regions that are smaller/harder to detect than holistic objects (\cf FSOD \cite{fan2020few}). Secondly, as shown in Fig.~\ref{fig:fskd-salvit}(a), the noisy patches from foreground and background will affect FSKD. Moreover, the predicted and groundtruth keypoints inherently exhibit ambiguity in location. While modeling both localization and semantic uncertainty \cite{Lu_2022_CVPR} helps, the interference from ambiguous local patterns is hard to solve. As each keypoint has close semantic and spatial relationships with other foreground patches, \eg left-eye lying left to right-eye and above on shoulder, such relations require ``context  modeling''. Thus, we argue that the long-range foreground relations may improve FSKD.

Vision transformer (ViT) \cite{dosovitskiy2020image, naseer2021intriguing} is a natural choice to capture such long-range relations. However, the global receptive field of self-attention (SA) in ViT leads to costly computations, and introduces irrelevant features outside of foregrounds \cite{xia2022vision}. Thus, we propose to use saliency maps, as they are convenient priors locating foreground objects that could help SA focus on foregrounds (Fig. \ref{fig:fskd-salvit}(b)) and ignore backgrounds. Interestingly, we also observe high similarity of  attentive regions of DINO \cite{caron2021emerging}  with saliency, which enables the use of DINO as both backbone and ``saliency'' detector to cut the number of parameters/computations.

In this paper, we present a saliency-guided vision transformer (SalViT) for few-shot keypoint detection (FSKD). Our SalViT is a plug-and-play module within encoder, which processes the raw support and query features individually, guided by saliency maps of support and query images. SalViT produces informative features relevant for support keypoint prototype (SKP) building, similarity learning, and query keypoint localization. Moreover, since the low-shot samples are insufficient to provide representative SKPs (\eg, 1-shot setting), we propose a transductive FSKD setting to enrich SKPs with unlabeled data. Finally, to improve robustness of our FSKD, we investigate masking and alignment (MAA) strategies, and probe the FSKD performance under occlusion setting.

\vspace{0.2cm}
Our contributions are summarized as follows:
% 1) We explore the utility of saliency map in vision transformer, use the saliency map as soft mask and propose a novel masked self-attention. With the masked self-attention (capture foreground semantic relevance/dependency).
% 2) We propose a morphology changing module within the vision transformer, which could automatically/dynamically learn the receptive field of SA.
% 3) With the masked self-attention, saliency map, and morphology learning module, we present the saliency-guided vision transformer. Further, we present cascade SalViT, and explore the fusion strategies of SalViT features, and use it to complement CNN feature, using both local and global features for FSL. (For this, we investigate using all ViT features; last ViT feature; last+middle. The preliminary results show that using all is better than simply using last one.)
% 4) We integrate our novel SalViT into pipelines of FSKD (and standard FSL in suppl.), validating the effectiveness of SalViT. In addition to inductive setting \cite{Lu_2022_CVPR}, we introduce a transductive algorithm for FSKD, which could further enhance performance.
\renewcommand{\labelenumi}{\roman{enumi}.}
\vspace{-0.1cm}
\begin{enumerate}%[leftmargin=0.6cm]
  \item We present a new variant of ViT called as saliency-guided vision transformer (SalViT), which has two key components: masked self-attention (M-SA) and morphology learner (ML). M-SA  captures distant relation between salient patches while ML adapts saliency and provides ``dynamic receptive field'' to exploit useful information in the proximity of foregrounds.\vspace{-0.2cm}
  \item We propose the pipeline of few-shot keypoint detection (FSKD) with SalViT. %, which benefits FSKD by unifying merits of modeling local and global dependency by fusing CNN and SalViT features.
  Moreover, a transductive extension of FSKD is explored. \vspace{-0.2cm}
  \item We propose a simple masking and alignment (MAA) strategy that helps detect occluded keypoints.
\end{enumerate}

Experiments on \emph{five} datasets show the superiority of the proposed SalViT-based FSKD model and the occlusion setting indicates that the simple MAA endows our model with stronger occlusion handling capability.

\section{Related Work}\label{sec:related-work}
% \subsection{Few-shot Learning (FSL) \& Tasks}
\noindent\textbf{Few-shot Learning (FSL):}
% 1) general few-shot learning and some representative works
% 2) computer vision tasks regarding few-shot learning (or few-shot tasks), standard FSL, FSOD, FSS, FSKD
% 3) the difference between our work and the existing works
%The goal of few-shot learning (FSL) is to infer the label of query image given one or a few support samples \cite{fei2006one,wang2020generalizing,koch2015siamese,vinyals2016matching}.
%Different from normal fully-supervised setting, the object classes of FSL in testing are unseen during training, which challenges the matching between query-support pairs. To tackle this issue,
Family of deep metric FSL models encourages the inter-class distance to be larger than the intra-class distance \cite{koch2015siamese,vinyals2016matching,snell2017prototypical,sung2018learning}. Vinyals \etal \cite{vinyals2016matching} formalize $N$-way $K$-shot protocol with episodes. Snell \etal \cite{snell2017prototypical} propose ProtoNet with centroids representing classes. %Dynamic weight generator (WG) methods \cite{gidaris2018dynamic,qiao2018few} use statistics of support features and the weight vectors of base classes to construct classifier.
 Sung \etal \cite{sung2018learning} propose a relation head to measure similarity. Other works emphasize adaptation and optimization (\eg MAML \cite{finn2017model}) and data hallucination \cite{zhang2019few}. %which increases the data variability in FSL. %and thus improve model generalization.

\noindent\textbf{Vision Transformers:}
% 1) transformer in NLP
% 2) vision transformer and its ingredients (components)
% 3) better architecture design, from the architecture specific to image classification to general-purpose architecture (introduce handcrafted designs for general-purpose vision tasks, cutting down computation complexity, SA)
% 4) better training strategy
% 5) analyze its strength and the key reason to success (majorly attributed to MSA)
%% Transformer has achieved tremendous success of sequence modeling in natural language processing (\eg machine translation). The difference between vision and language domains stems from the information structure: 2D spatial array (pixels) for image data  \vs 1D array (words) for language text. Transformer--adaptation to--vision transformer.
Transformers \cite{vaswani2017attention} are popular in sequential modeling/natural language processing (\eg, BERT \cite{devlin2018bert} and GPT-3 \cite{brown2020language}), %. They have also been adopted in
and computer vision  \cite{parmar2018image}. Vision Transformer (ViT) \cite{dosovitskiy2020image} has inspired data efficient DeiT \cite{touvron2021training}, self-supervised DINO \cite{caron2021emerging}, and CrossViT \cite{chen2021crossvit}, \etc. Although ViTs often outperform CNNs on classification problems, they do not provide multi-scale features for dense prediction tasks. Thus, other general-purpose architectures such as Swin-T \cite{liu2021swin} and Twins \cite{chu2021twins} were proposed. %, yet involving more inductive bias. On top of vanilla ViT, we delve into designing a ViT guided by visual saliency for FSKD.

\noindent\textbf{Masked Attentions:}
% Attention mechanism enhances/complements CNNs
% 1) vanilla ViT mask strategy in SA and our more in-depth ones
% 2) SA layer complements CNN (Embedded in network case: channel attention in Squeeze-and-Excitation Networks, or spatial attention; Appended case: Attention Aug. (closely related to our work), etc.)
% 3) Explore the effects of ViT feature in FSL
% Note: attention in Transformer/CNN; attention in FSL (Cross attention network for few-shot classification)
%%\noindent\textbf{Self-attention/Transformers to complement CNNs:} selft-attention: interaction between tokens, long-range dependency, weighted sum to attend to most informative tokens. Self-attention can also be regarded as information spreading, where the patch tokens spread information to class token in inference while the supervision information spreads from class token to patch tokens \cite{touvron2021training}.
Attention mechanisms include \emph{self-attention} (SA) and \emph{cross-attention} (CA) \cite{vaswani2017attention,hou2019cross}. SA is crucial in ViT modeling but its computation is high as SA operates globally. To remedy this, Swin-T \cite{liu2021swin} applies SA within the masks of
local windows. Twins \cite{chu2021twins} perform locally-grouped self-attention and then global sub-sampled attention. Unlike Swin-T and Twins using regular windows (\ie, rectangle) with hard masking, we use saliency map to obtain ``soft masks'' of foreground shapes, serving as the foreground attention. The masked attention is widely used \cite{cai2022mask,cheng2022masked,li2022mat} for invalid-token masking, self-supervised training \cite{he2022masked,xie2022simmim}, image inpainting \cite{li2022mat}, \etc. However, the masked self-attention (M-SA) in our SalViT is uniquely crafted for FSKD. Our M-SA cooperates with a morphology learner which provides ``dynamic receptive field'' for adaptation on top of detected salient regions.

\begin{figure*}[!t]
  \vspace{-10pt}
  \centering
  \includegraphics[width=.9\linewidth]{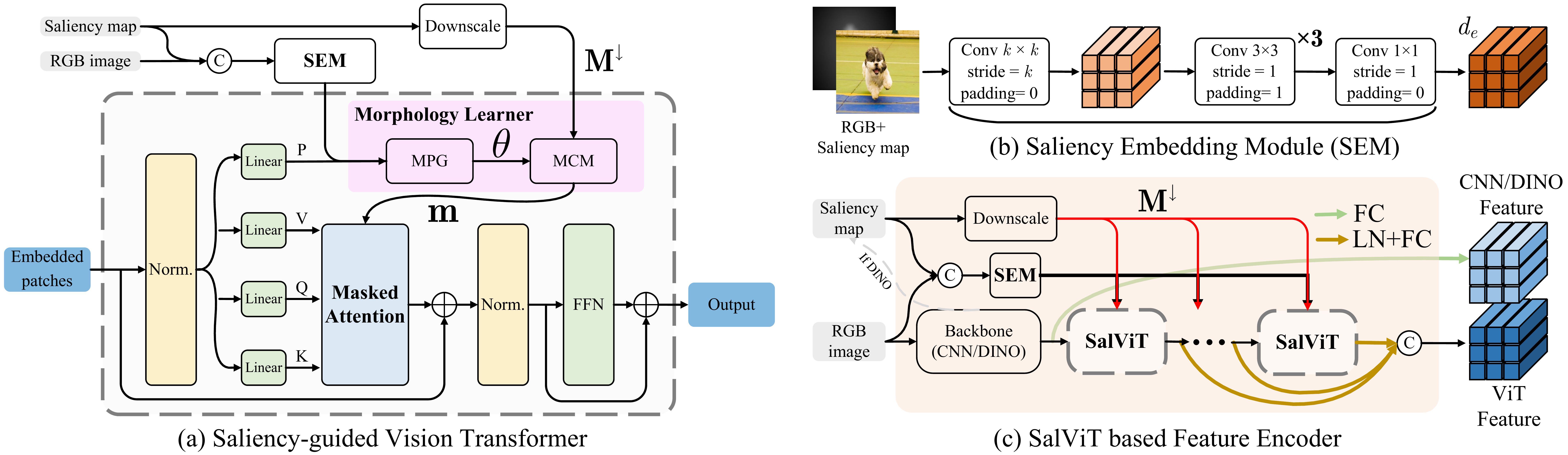}
  \caption{Model overview. (a) The architecture of saliency-guided vision transformer (SalViT); (b) details of semantic embedding module (SEM); (c) SalViT-based feature encoder $\mathcal{E}$ for FSKD.
  }
  \label{fig:salvit-structure}
  \vspace{-10pt}
\end{figure*}

\section{Method}\label{sec:approach}
% In this section, we will first briefly revisit vanilla ViT \cite{dosovitskiy2020image}, then present our saliency-guided vision transformer (SalViT), and finally showcase the few-shot keypoint detection (FSKD) with SalViT.
\subsection{Preliminary}
% 1). vanilla ViT: ingredients include Q,K,V projection layers, self-attention, ffn, layer norm.
%
%Notations are in Section \ref{app:notations}.
%We first revisit
\noindent
\textbf{Vanilla ViT \cite{dosovitskiy2020image}.} For a sequence of $n$ tokens  of $d$ dimensional feature vectors in $\mathbf{X} \in\mathbb{R}^{n \times d}$, $\mathbf{X}$ forms $\mathbf{Q}$, $\mathbf{K}$, $\mathbf{V} \in\mathbb{R}^{n \times d}$ (\ie query, key, and value) by linear projections
\begin{equation}
  \mathbf{Q} = \mathbf{X}\mathbf{W}_{q}, \quad \mathbf{K} = \mathbf{X}\mathbf{W}_{k}, \quad \mathbf{V} = \mathbf{X}\mathbf{W}_{v},
\end{equation}
where $\mathbf{W}_{q}$, $\mathbf{W}_{k}$, and $\mathbf{W}_{v}$ are projection matrices. The self-attention (SA) is applied to learn relations between tokens %, which can be formulated as
\begin{equation}
    %\tilde{V}_{i,:} = \sum\nolimits_j \phi(Q_{i,:}, K_{j,:}) \odot V_{j,:} = \sum\nolimits_j A_{i,j} \odot V_{j,:}, \quad \tilde{z} = \tilde{V}W_{o},
    A_{i,j} = \gamma(\phi(\mathbf{Q}_{i,:}, \mathbf{K}_{j,:})), \quad \tilde{\mathbf{V}} = \mathbf{A}\mathbf{V}, \quad \mathbf{Z} = \tilde{\mathbf{V}}\mathbf{W}_{o},
    \label{eq:vanilla-SA}
\end{equation}
where $\gamma(\cdot)$ is a non-linear function and $\phi(\cdot)$ is a similarity function, the two together yielding attention $\mathbf{A} \in \mathbb{R}^{n \times n}$. $\mathbf{W}_o$ is a projection matrix, and $\mathbf{Z}$ is the output of SA. Functions $(\gamma, \phi)$ can be the \emph{scaled dot-product based softmax SA}, \ie, $\gamma(\cdot)=\text{softmax}(\cdot)$, $\phi(\mathbf{Q}_{i,:}, \mathbf{K}_{j,:})=\mathbf{Q}_{i,:}\mathbf{K}_{j,:}^{\intercal}/(\beta\sqrt{d})$, or \emph{Gaussian kernel based RBF SA}, \ie, $\gamma(\cdot)=\exp(\cdot)$, $\phi(\mathbf{Q}_{i,:}, \mathbf{K}_{j,:})=-\|\mathbf{Q}_{i,:} - \mathbf{K}_{j,:}\|/(2\beta\sqrt{d})$.
%$\phi$ can be the scaled dot-product based softmax SA, \ie $\phi(Q, K)=\text{softmax}(QK^{\intercal}/\sqrt{d})$, or Gaussian kernel based RBF SA, \ie $\phi(Q_{i,:}, K_{j,:})=\exp(-\|Q_{i,:} - K_{j,:}\|/(2\beta\sqrt{d}))$.
%To avoid the numeric instability when $d$ gets large and
To accelerate computation, $\mathbf{Q}$, $\mathbf{K}$, $\mathbf{V}$ are split into multiple sub-matrices in channel dimension and computed via multi-head self-attention (MHSA). Finally, a feedforward network (FFN) comprised by two linear layers with GELU activation is used. % to provide non-linearity.
With the layer normalization (LN) and skip connection, a basic ViT block becomes
\begin{equation}
  \mathbf{Z}'_{l} = \text{MHSA}(\text{LN}(\mathbf{Z}_{l-1})) + \mathbf{Z}_{l-1},\!\!\quad \!\!\mathbf{Z}_{l} = \text{FFN}(\text{LN}(\mathbf{Z}'_{l})) + \mathbf{Z}'_{l},
\end{equation}
where $\mathbf{Z}_{l}$ is the output of $l$-th ViT. Based on vanilla ViT, we will build our SalViT in next section.

% SA aggregates patch features (or ``tokens'') globally via a linear weighted summation.

\subsection{Saliency-guided Vision Transformer}
% 1) vanilla ViT: ingredients include Q,K,V projection layers, self-attention, ffn, layer norm.
% 2) show our salvit structure, for sake of comparing vanilla ViT

Our SalViT is shown in Fig.~\ref{fig:salvit-structure}(a), which includes masked self-attention (M-SA), morphology learner (ML), and saliency embedding module (SEM), in addition to the components of  %normal LN and FFN in
vanilla ViT. The ML takes patch features $\mathbf{P}$ and saliency embeddings produced by SEM as input, and predicts a conditional parameter $\theta$ to change the morphology of saliency map, yielding an adapted mask for M-SA. In what follows, we will detail each component of SalViT. % and the mechanism of patching relation learning with our M-SA that is beneficial to FSKD.

%The overview of our SalViT is showed in Fig.~\ref{fig:salvit-structure}(a), which includes masked self-attention (M-SA), morphology learner, and saliency embedding module (SEM), in addition to the normal LN and FFN in vanilla ViT. The saliency embeddings $F_{sal}$ produced by SEM is sharable for all SalViTs as they stem from same image. The morphology learner consists of a morphology parameter generator (MPG) and a morphology changing module (MCM). MPG takes both patch features $P$ and $F_{sal}$ as input and predicts conditional parameter $\theta$ for MCM to change the morphology of saliency map, yielding an adapted mask $M$ for M-SA. In the following, we will detail each component and the mechanism of patching relation learning with our M-SA, which is beneficial to FSKD.

\subsubsection{Saliency Map Generation}
% 1) introduce saliency detectors: conventional detectors (unsupervised), deep detectors (weakly supervised, and supervised).
% 2) show some images for comparisons + analysis of potential infulence to FSL (worse pck for unsupervised det./better for deep detector, etc.)
% 3) pre-image processing: distance transform and gaussian blur
% note that the SCRN/MNL model is pre-trained on DUTS, which has no intersection with FSKD datasets.
\begin{table}
  \centering
  \caption{Statistic mean-IoU of each saliency prior to that of SCRN.}\label{tab:ious}
  \setlength{\tabcolsep}{2.5pt}
  \begin{tabular}{c|cccc}
    mIoUs         & MNL & DINO & RBD & WScribble\\ \specialrule{0.1em}{0em}{0em}
    Animal Dataset \cite{cao2019cross} &0.698&0.576 &0.343& 0.757    \\
    AwA Dataset \cite{banik2021novel}   &0.697&0.575 &0.412& 0.753    \\
  \end{tabular}
  \vspace{-0.3cm}
\end{table}
Visual saliency, relating to human visual attention, is given as a single-channel map with values in $[0, 1]$ range. The conventional saliency detectors (\eg, RBD \cite{zhu2014saliency}) are modeled by various handcrafted priors, and underperform on complex scenes. More robust deep saliency detectors  include i) unsupervised detectors, \eg MNL \cite{zhang2018deep}; ii) supervised ones, \eg SCRN \cite{wu2019stacked} and WScribble \cite{zhang2020weakly}. MNL is robust and it  does not use human annotations \cite{zhang2018deep,zhang2019few}. %\footnote{Saliency dataset for training sal. is disjoint with FSKD datasets.}.

We validate the performance of each saliency detector by computing mean IoU to the state-of-the-art SCRN on two FSKD datasets (Table \ref{tab:ious}). Also, inspired by the self-supervised vision transformer DINO \cite{caron2021emerging}, and its impressive unsupervised semantic segmentation, we treat the \emph{class token attentiveness} \cite{liang2021evit,caron2021emerging} from the pre-trained DINO (ViT-S/16) as another form of ``saliency prior''. DINO yields over 57\% mean IoU: it captures salient regions. Thus, we use unsupervised MNL or DINO for saliency priors in  main experiments. We investigate %the performance of different
other saliency detectors in Sec.~\ref{sec:ablation_study}. With saliency, ViT captures salient patch relations well.

\subsubsection{Soft Masked Self-attention}
% vanilla self-attention (SA)
% assume we have a saliency map as soft mask M, then introduce soft masked self-attention
% - saliency interation matrix
% - masking as bias in SA
% - relative position encoding
% - remarks: extreme cases of masked SA, no masking (FC) vs. all masking (vanilla SA)
The vanilla SA is given as $A_{i,j} = \gamma(\phi(\mathbf{Q}_{i,:}, \mathbf{K}_{j,:}))$ (Eq. \ref{eq:vanilla-SA}). We take downscaled saliency map $\mathbf{M}^{\downarrow}$ flattened as $\mathbf{m} \in \mathbb{R}^{n}$ ($n$ is the number of tokens) to incorporate saliency into SA as $A_{i,j} = \gamma(\psi(\mathbf{Q}_{i,:}, \mathbf{K}_{j,:}, m_i, m_j))$, where $m_i$, $m_j$ are the saliency of $i$-th and $j$-th patches, respectively. Intuitively, the attention should be encouraged to attend patch relations on target object  given $\mathbf{m}$. We define ($\gamma$, $\psi$) as
\begin{equation}
  % \begin{aligned}
    \gamma(\psi(\mathbf{Q}_{i,:}, \mathbf{K}_{j,:}, m_i, m_j)) \!=\!
    \left\{\!\!
        \begin{array}{cr}
            \gamma(\phi(\mathbf{Q}_{i,:}, \mathbf{K}_{j,:})) &\kern-2.3em i,j\in \mathcal{S}_{+}  \\
            0           & \kern-2.3em   i \in \mathcal{S}_{+}, j \in \mathcal{S}_{-}  \\
            \delta(i-j) & \kern-2.3em   i \in \mathcal{S}_{-}
        \end{array}
    \right.
  % \end{aligned}
  \label{eq:hard-masking}
\end{equation}
where $\mathcal{S}_{+}\!\equiv\!\{i\!:\! m_i\!\geq\!0.5\}$ (or $\mathcal{S}_{-}\!\equiv\!\{i\!:\! m_i\!<\!0.5\}$) stands for the index set of foreground (or background) patches and $\delta(i-j)$ is 1 if $i=j$, 0 otherwise. Our choice of ($\gamma$, $\psi$) means that the relation of the pair of foreground patches is learned, foreground-background pair is inactive, and background patch only focuses self-information. However, such hard masking is inapplicable in real scenario as one patch grid may mix both foreground and background information. Thus, we propose a soft-masked self-attention (M-SA). Firstly, we define a so-called saliency interaction matrix (SIM), $\bar{\mathbf{M}}\in \mathbb{R}^{n\times n}$, which represents the saliency degree of paired patches. We can use 1) dot-product, \ie $\bar{\mathbf{M}}=\mathbf{m}\mathbf{m}^{\intercal}$; 2) harmonic mean, \ie $\bar{\mathbf{M}}=2\mathbf{m}\mathbf{m}^{\intercal}/(\mathbf{m}\mathbf{1}^\intercal+\mathbf{1}\mathbf{m}^{\intercal}+\epsilon)$ ($\epsilon\rightarrow 0$); and 3) arithmetic mean, \ie $\bar{\mathbf{M}}=(\mathbf{m}\mathbf{1}^\intercal+\mathbf{1}\mathbf{m}^{\intercal})/2$, \etc. We select \emph{harmonic mean} by default as it performed better in our experiments. In SIM, the background patch focusing on itself is not included. Thus, a term $\mathbb{I}-\text{Diag}(\mathbf{m})$ is added to form the attention mask $\tilde{\mathbf{M}} \in \mathbb{R}^{n \times n}$ as
\begin{equation}
    \tilde{\mathbf{M}}=\bar{\mathbf{M}} + \mathbb{I}-\text{Diag}(\mathbf{m}),
    \label{eq:attention-mask}
\end{equation}
where $\text{Diag}(\mathbf{m}) \in \mathbb{R}^{n \times n}$ is the diagonal matrix of $\mathbf{m}$. A property of $\tilde{\mathbf{M}}$ is the diagonal elements are all \emph{unit}. %, which circumvent the pitfall of SIM.
By treating $\tilde{\mathbf{M}}$ as a bias, we formulate soft M-SA as
\begin{equation}
    % \gamma(\psi(Q_{i,:}, K_{j,:}, M_i, M_j)) = \gamma(\phi(Q_{i,:}, K_{j,:}) - (1-\tilde{M}_{i,j})J),
    \mathbf{A} = \gamma(\psi(\mathbf{Q}, \mathbf{K}, \mathbf{M})) = \gamma(\phi(\mathbf{Q}, \mathbf{K}) - (\mathbf{1}\mathbf{1}^\intercal-\tilde{\mathbf{M}})J),
    \label{eq:soft-masking}
\end{equation}
where $J$ is a scalar that controls the degree of masking, and ($\gamma$, $\phi$) is the softmax SA or RBF SA. In contrast to hard masking in Eq.~\ref{eq:hard-masking}, one can back-propagate through Eq. \ref{eq:soft-masking} of soft M-SA. Also, compared to the global attention of vanilla SA (Eq.~\ref{eq:vanilla-SA}), our M-SA facilitates  relation learning in salient areas.

\vspace{0.1cm}
\noindent\textbf{Property of M-SA:} If the whole image is salient, \ie, $\mathbf{m}=\mathbf{1} \in \mathbb{R}^{n}$, then $\tilde{\mathbf{M}}=\bar{\mathbf{M}}=\mathbf{1}\mathbf{1}^\intercal \in \mathbb{R}^{n \times n}$. Thus,
% $\gamma(\psi(Q_{i,:}, K_{j,:}, M_i, M_j))=\gamma(\phi(Q_{i,:}, K_{j,:}))$
$\gamma(\psi(\mathbf{Q}, \mathbf{K}, \mathbf{M}))=\gamma(\phi(\mathbf{Q}, \mathbf{K}))$
which recovers the vanilla SA. % which learns relations on full image.
If the whole image is non-salient, \ie, $\mathbf{m}=\mathbf{0}$, then $\bar{\mathbf{M}}=\mathbf{0}$, $\tilde{\mathbf{M}}=\mathbb{I}$ then
% $\gamma(\psi(Q_{i,:}, K_{j,:}, M_i, M_j))=\gamma(\phi(Q_{i,:}, K_{j,:}) - (1-I_{i,j})J)$
$\gamma(\psi(\mathbf{Q}, \mathbf{K}, \mathbf{M}))=\gamma(\phi(\mathbf{Q}, \mathbf{K}) - (\mathbf{1}\mathbf{1}^\intercal-\mathbb{I})J)$
which means that all attention entries are masked except for the diagonal (M-SA learns no relations). %The above are two extreme cases of behavior of M-SA.

\subsubsection{Morphology Learner}
% 1. modules include:
% - saliency embedding module (SEM)
% - morphology parameter generator
% - morphology changing module
% 2. regularization loss (optional)
\begin{figure}[!t]
  \centering
  \includegraphics[width=\linewidth]{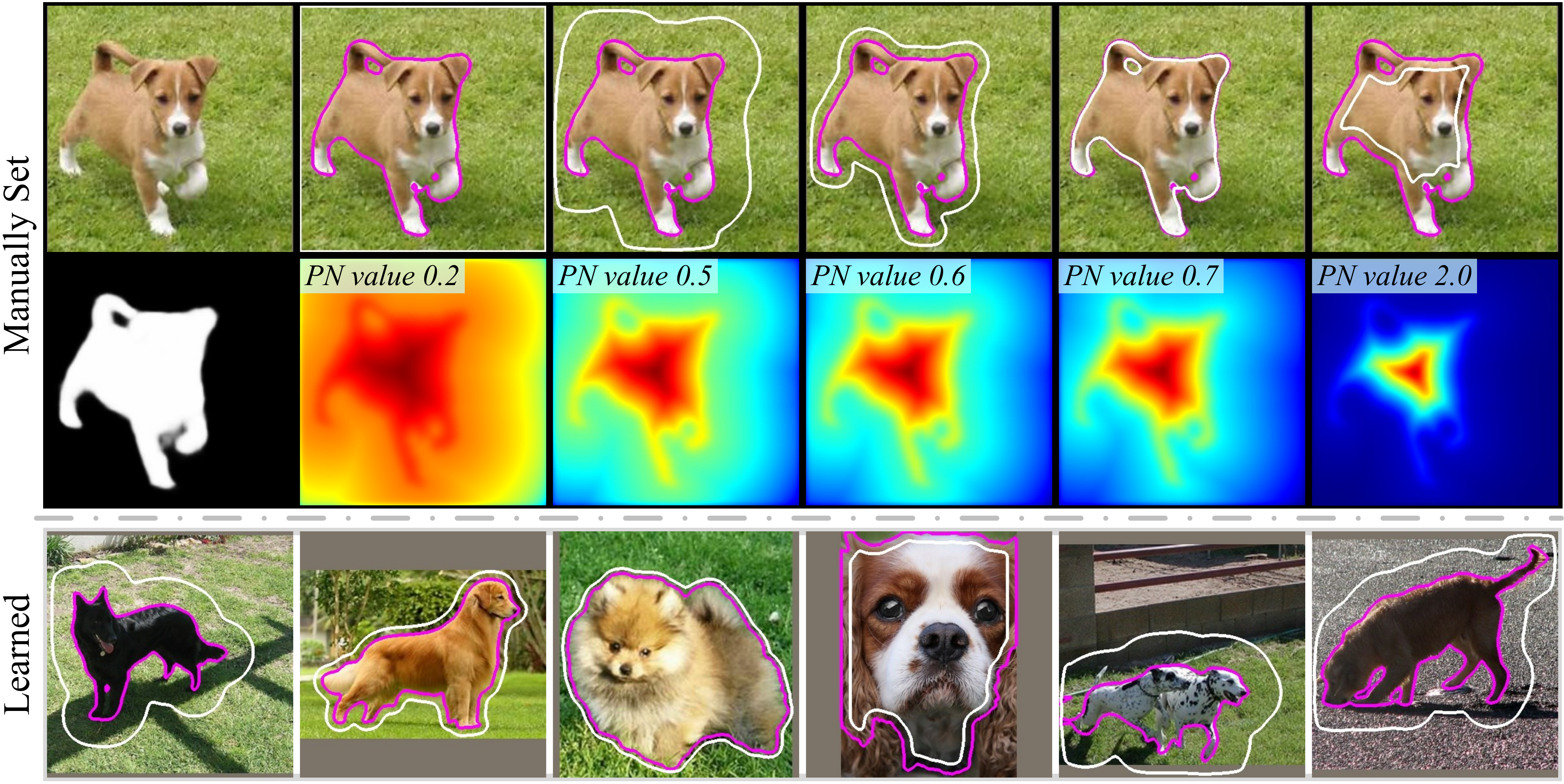}
  \caption{The illustration of power normalization (PN). The top images show the effects when PN values are \emph{mannually set} while the bottom images show the morphology \emph{learned} by our ML.
  }
  \label{fig:power-norm}
\end{figure}

Although M-SA can directly use the downscaled saliency map $\mathbf{M}^{\downarrow}$ , it lacks adaptation to the context of object. To endow SalViT with a ``dynamic receptive field'', we propose the morphology learner (ML) which modifies $\mathbf{M}^{\downarrow}$. As shown in Fig.~\ref{fig:salvit-structure}(a), the morphology learner consists of a morphology parameter generator (MPG) and a morphology changing module (MCM). Firstly, we adopt a saliency embedding module (SEM) to encode saliency map and RGB image into saliency embeddings $\mathbf{F}^{\text{sal}} \in \mathbb{R}^{n \times d_e}$ (Fig.~\ref{fig:salvit-structure}(b)). Then, the patch features $\mathbf{P}$ and $\mathbf{F}^{\text{sal}}$ are concatenated and globally average-pooled as $\mathbf{F}=\text{GAP}([\mathbf{F}^{\text{sal}}, \mathbf{P}]) \in \mathbb{R}^{(d+d_e)}$ and fed to MPG to predict parameter $\theta$ as
% \vspace{-0.2cm}
\begin{equation}
    \theta = \mathbf{W}_{2}\texttt{GeLU}(\mathbf{W}_{1}\mathbf{F} + \mathbf{b}_{1}) + \mathbf{b}_{2}.
\end{equation}
Finally, $\theta$ is injected into MCM to control the morphology of $\mathbf{M}^{\downarrow}$ via \emph{power normalization} as
% \vspace{-0.1cm}
\begin{equation}
    \tilde{\theta} = \rho_1 \cdot \text{sigmoid}(\theta), \quad \mathbf{M} = (\mathbf{M}^{\downarrow})^{\tilde{\theta}},
\end{equation}
where $\rho_1 \in \mathbb{R}$ controls the bound of morphology ($0 \leq \tilde{\theta} \leq \rho_1$). To ensure $\tilde{\theta}$ changes properly, we regularize it by  $\mathcal{L}_{\text{reg}}=\mathbb{E}[\max((\tilde{\theta}-\rho_2)^2-\rho_3, 0)]$, where $\rho_2$ is the desired morphology center and $\rho_3$ is the interval. We set $\rho_1=2$, $\rho_2=0.7$, and $\rho_3=0.05$. Thus, matrix $\mathbf{M}$ (or $\mathbf{m}$ in vectorized form) is learned  to help M-SA focus on foregrounds and the context (Fig.~\ref{fig:power-norm}).

\subsection{Few-shot Keypoint Detection}
\begin{figure*}[!t]
\vspace{-0.3cm}
    \centering
    \includegraphics[width=.90\linewidth]{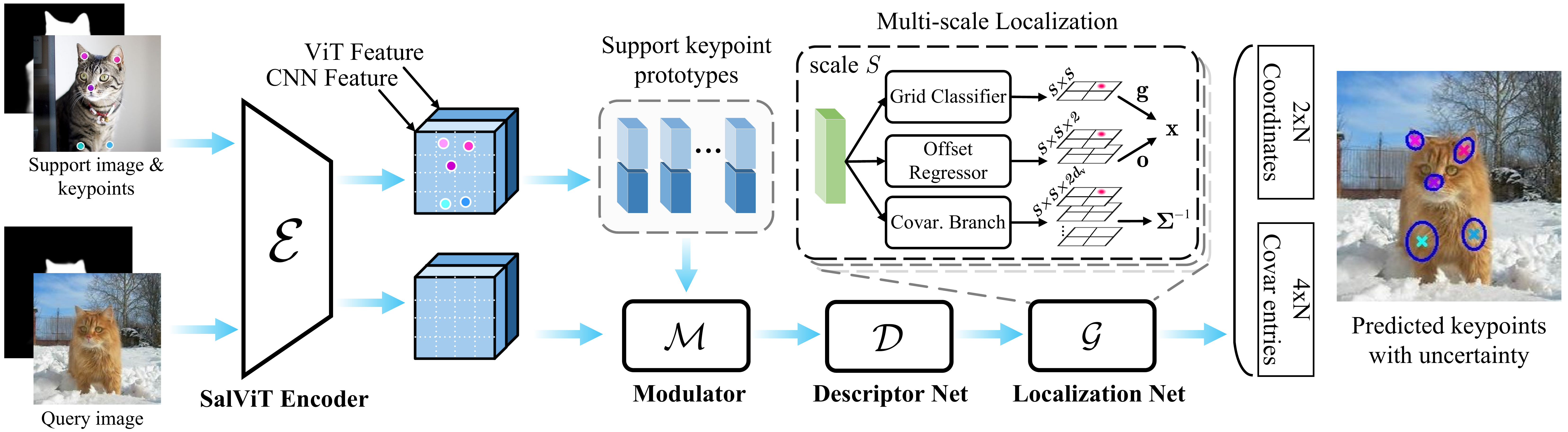}
    \caption{Overview of the proposed few-shot keypoint detection pipeline with SalViT.
    }
    \label{fig:pipeline-fskd-with-salvit}
\end{figure*}

% pipeline of FSKD: encoder with salvit, feature modulator, descriptor extractor, keypoint localization net with uncertainty estimation (uncertainty-aided grid-based locator (UC-GBL), Multi-scale UC-GBL and uncertainty fusion)
% loss functions
% In this section, the SalViT will be integrated into our few-shot keypoint detection (FSKD) pipeline, as shown in Fig.~\ref{fig:pipeline-fskd-with-salvit}.
Based on SalViT, we propose our few-shot keypoint detection (FSKD) pipeline, as shown in Fig.~\ref{fig:pipeline-fskd-with-salvit}. For $K$ support images and $N$ support keypoints per image,  FSKD detects the corresponding keypoints in a query image (\emph{$N$-way-$K$-shot detection}). Our FSKD model includes SalViT-based feature encoder $\mathcal{E}$, feature modulator $\mathcal{M}$, descriptor net $\mathcal{D}$, and keypoint localization net $\mathcal{G}$. Features of $\mathcal{E}$ capture relations between salient patches, which helps modulator $\mathcal{M}$.

%\noindent\textbf{Feature Augmentation with SalViT:} % cascade SalViT complements CNN features
% show CNN + Cascade SalViT backbone for FSL (need a figure to show, versitility of SalViT)
% feature selection and fusion for salvit features from different layers
% CNN feature + fused salvit feature for FSL
Fig.~\ref{fig:salvit-structure}(c) shows the architecture of $\mathcal{E}$, which takes saliency/RGB $\mathbf{I}\!=\![\mathbf{I}^{\text{sal}}; \mathbf{I}^{\text{rgb}}] $ as inputs, and outputs ensemble features $\mathcal{E}(\mathbf{I})$ that encode foreground patch relations. %which is the concatenation of token features and ViT features. Thus, $\mathcal{E}$ contains local and global features.

\comment{
Fig.~\ref{fig:salvit-structure}(c) shows the architecture of $\mathcal{E}$, which takes saliency/RGB images $\mathbf{I}\!=\![\mathbf{I}^{\text{sal}}; \mathbf{I}^{\text{rgb}}] $ as inputs, and outputs ensemble features  $\mathcal{E}(\mathbf{I})\!=\![\mathbf{F}^{\text{cnn}}; \mathbf{F}^{\text{vit}}] \!\in\! \mathbb{R}^{l \times l \times d}$ where $d\!=\!d_{\text{cnn}}\!+\!d_{\text{vit}}$. Thus, $\mathcal{E}$ contains local and global features.
}

\vspace{0.1cm}
\noindent\textbf{Inductive Inference:}
% (remember to use steps, little break or disruption)
Let the $k$-th support image with saliency $\mathbf{I}^\text{s}_{k}$ be encoded as $\mathcal{E}(\mathbf{I}^\text{s}_{k}) \in \mathbb{R}^{l \times l \times d}$, and $\mathbf{X}^\text{s}_{k,n}$ is the $n$-th support keypoint in the $k$-th support image. Firstly, we extract support keypoint representation (SKR) $\mathbf{\Phi}_{k,n}\in\mathbb{R}^{d}$ via Gaussian pooling. %\ie ${\mathbf{\Phi}}_{k, n} = \sum\nolimits_{\mathbf{p}} \exp(-\parallel \mathbf{p}-{\mathbf{X}}^\text{s}_{k,n} \parallel _{2}^{2}/2\xi  ^2) \cdot \mathcal{E}(\mathbf{I}^\text{s}_{k})(\mathbf{p})$, where $\mathcal{E}(\mathbf{I}^\text{s}_{k})(\mathbf{p})$ means the feature vector at position $\mathbf{p} \in \mathbb{R}^2$.
Following PrototNet \cite{snell2017prototypical}, we form the support keypoint prototype (SKP) to represent each type of keypoints by averaging the SKRs across support images, \ie, ${\mathbf{c}}_n = \frac{1}{K} \sum\nolimits_{k=1}^{K} {\mathbf{\Phi}}_{k, n}$. Thus, $N$ SKPs are produced for $N$-way-$K$-shot detection. Subsequently, the SKPs are used to modulate query feature map $\mathcal{E}(\mathbf{I}^\text{q}) \in \mathbb{R}^{l \times l \times d}$ via $\mathcal{M}$, yielding the attentive feature map $\mathbf{F}^{\text{att}}_n=\mathcal{M}(\mathcal{E}(\mathbf{I}^\text{q}), {\mathbf{c}}_n) \in \mathbb{R}^{l \times l \times d}$ per SKP, %Concretely, $\mathcal{M}$ could have various choices such as simple correlation, spatial attention, channel attention, \etc. We choose $\mathcal{M}$ as simple correlation due to its efficiency, namely $F_{\text{att}}^n(\mathbf{p})=\mathcal{E}(I^\text{q})(\mathbf{p}) \odot {\mathbf{c}}_n$.
where $\mathbf{F}^{\text{att}}_n(\mathbf{p})=\mathcal{E}(\mathbf{I}^\text{q})(\mathbf{p}) \odot {\mathbf{c}}_n$ and $\odot$ refers channel-wise multiplication. Thus $\mathcal{M}$ %is to perform similarity learning in order to
correlates SKPs with similar regions in the query image. % to localize query keypoints.
Each $\mathbf{F}^{\text{att}}_n$ is then processed into descriptor $\mathbf{\Psi}_n$ via $\mathcal{D}$ %that is composed by three conv. layers.
and fed to a multi-scale grid-based localization net $\mathcal{G}$ to predict keypoint $\mathbf{x} \in \mathbb{R}^2$ and uncertainty $\mathbf{\Sigma} \in\mathbb{R}^{2\times 2}$.

Fig.~\ref{fig:pipeline-fskd-with-salvit} shows that each scale of localization net $\mathcal{G}^{(S)}$ includes three branches: grid classifier $\mathcal{G}_\text{c}$, offset regressor $\mathcal{G}_\text{o}$, and covariance branch $\mathcal{G}_\text{v}$, which  predict grid probability map $\mathbf{P}\!=\!\text{softmax}(\mathcal{G}_\text{c}(\mathbf{\Psi}_n)) \in \mathbb{R}^{S \times S}$, offset vector field $\mathbf{O}=\mathcal{G}_\text{o}(\mathbf{\Psi}_n) \in \mathbb{R}^{S \times S \times 2}$, and latent covariance field $\mathbf{V}=\mathcal{G}_\text{v}(\mathbf{\Psi}_n)) \in \mathbb{R}^{S \times S \times 2d_{\text{v}}}$, respectively. The grid $\mathbf{g} \in \{0, \!\cdots\!, S\!-\!1\} \times \{0, \!\cdots\!, S\!-\!1\}$. % that has maximal score is indexed from $\mathbf{P}$.
Accordingly, we extract offset $\mathbf{o}=O(\mathbf{g}) \in [-1, 1)^2$ and latent covariance $\mathbf{Q}=\mathbf{V}(\mathbf{g}) \in \mathbb{R}^{2d_v}$ using $\mathbf{g}$. In this paper, we regress precision matrix $\mathbf{\Omega}=\tilde{\mathbf{Q}}(\tilde{\mathbf{Q}})^{\intercal}/d_{v}$ instead of covariance $\mathbf{\Sigma}$ to avoid inverting covariance in the loss function, where $\tilde{\mathbf{Q}} \in \mathbb{R}^{2 \times d_v}$ is reshaped from $\mathbf{Q}$. We form the grid classification loss $\mathcal{L}_{\text{cls}}$ and offset regression loss $\mathcal{L}_{\text{os}}$ as
\vspace{-0.2cm}
\begin{equation}
  % \begin{aligned}
\!\!\!\!\!\mathcal{L}_{\text{cls}} = - \mathbb{E} \log \mathbf{P}(\hat{\mathbf{g}}),
    \mathcal{L}_{\text{os}} = \frac{1}{2}\mathbb{E}[(\mathbf{x} - \hat{\mathbf{x}})^{\intercal} {\mathbf{\Omega}} (\mathbf{x} - \hat{\mathbf{x}}) - \log\det({\mathbf{\Omega}})],
  % \end{aligned}
\end{equation}
where $\hat{\mathbf{g}}$ and $\hat{\mathbf{x}}$ are groundtruth grid and offset. If $\mathcal{G}$ has $N_S$ scales $\{S_i\}_{i=1}^{N_S}$, the multi-scale localization loss is $\mathcal{L}_{\text{ms}}=\sum\nolimits_{i=1}^{N_S}(\mathcal{L}_{\text{cls}}^{(S_i)} + \mathcal{L}_{\text{os}}^{(S_i)})/N_S$. Moreover, the multi-scale prediction is obtained as
\begin{equation}
  \begin{aligned}
    &\mathbf{x}=\frac{1}{N_S} \sum\nolimits_{i=1}^{N_S} \frac{l_{0}}{S_i}\Big(\mathbf{g}^{(S_i)}+0.5+0.5\mathbf{o}^{(S_i)}\Big) \\
    &\mathbf{\Sigma} = \frac{1}{4N_{S}}\sum\nolimits_{i=1}^{N_S} \Big(\frac{l_0}{S_i}\Big)^2 {\mathbf{\Sigma}}^{(S_i)}
  \end{aligned},
\end{equation}
where $l_0$ is the edge length of square-padded query image, ${\mathbf{\Sigma}}^{(S_i)}$ is the inverse of $\mathbf{\Omega}^{(S_i)}$ at scale $S_i$.

\vspace{0.1cm}
\noindent\textbf{Masking and Alignment (MAA):}
Even though ViT is known to be robust to occlusions \cite{naseer2021intriguing}, it  fails under severe occlusions over keypoints (see Table~\ref{tab:MAA-bgcrop}). Inspired by \cite{zhong2020random}, we propose to randomly mask input $\mathbf{I}=[\mathbf{I}^{\text{sal}}; \mathbf{I}^{\text{rgb}}]$ to form an occluded view $\mathbf{I}^{\text{occ}}$. We define localization loss on occluded inputs as $\mathcal{L}_{\text{ms}}^{\text{occ}}$. Then we form an alignment loss $\mathcal{L}_{\text{aln}}$ by the KL divergence $\text{KL}[\mathbf{P} || \mathbf{P}^{\text{occ}}]$ or $\|\mathcal{E}(\mathbf{I}) - \mathcal{E}(\mathbf{I}^{\text{occ}})\|$ with the $\ell_1$ or $\ell_2$ loss.  $\mathbf{P}$ (or $\mathbf{P}^{\text{occ}}$) and $\mathcal{E}(\mathbf{I})$ (or $\mathcal{E}(\mathbf{I}^{\text{occ}})$) refer to grid probability maps and features of non-occluded (or occluded) inputs. In experiments we investigate the best strategy to perform MAA. Including the regularization loss $\mathcal{L}_{\text{reg}}$ obtained from $\mathcal{E}$,  the total loss becomes
\begin{equation}
    \mathcal{L}=\lambda_{1}\mathcal{L}_{\text{ms}}^{\text{occ}} + \lambda_{2}\mathcal{L}_{\text{reg}} + \lambda_{3}\mathcal{L}_{\text{aln}}, %\quad\text{($\lambda_{1} = \lambda_{2} = 0.5, \lambda_{3}=0.1$)}.
    \label{eq:total-loss}
\end{equation}
where we set $\lambda_{1} = \lambda_{2} = 0.5$, $\lambda_{3}=0.1$ by default.

\vspace{0.1cm}
\noindent\textbf{Transductive Inference:}
% note that the key of transductive FSKD is to select high-quality pseudo labels
 FSKD has to represent prototype $\mathbf{c}_n$ for each kind of support keypoints well, and measure its similarity \wrt the query feature. As support samples are limited, SKP $\mathbf{c}_n$ cannot represent the true keypoint well. Thus, we extend FSKD to the transductive setting by leveraging the unlabeled query samples.

Let SKRs be $\mathcal{S}\!\equiv\!\{{\mathbf{\Phi}_{k,n}}\}_{k,n=1}^{K\times N}$ and the SKRs of $n$-th keypoint type be $\mathcal{S}^{(n)}\!\equiv\!\{{\mathbf{\Phi}_{k,n}}\}_{k=1}^{K}$. Let the initial SKPs $\mathcal{C}\!\equiv\!\{{\mathbf{c}_{n}}\}_{n=1}^{N}$ be formed based on $\mathcal{S}$. For each query image and each prototype, we retrieve top-$W$ query keypoints. For $Z$ query images, we get $\{\mathbf{x}_{z,n,w}\}_{z,n,w=1}^{Z\times N \times W}$, by selecting the top-$W$ grids from the grid probability map. Then we define the label confidence criterion by using the corresponding keypoint probability scores $\{p_{z,n,w}\}_{z,n,w=1}^{Z\times N\times W}$, rank the scores per keypoint type in descending order, and select top-$\eta$ query keypoints per type to generate the candidate set $\mathcal{Q}^{(n)}\!\equiv\!\{ \tilde{\mathbf{\Phi}}\! :\! \tilde{\mathbf{\Phi}}\!=\!\mathcal{E}(\mathbf{I}^{\text{q}}_{z})(\mathbf{x}_{z,n,w}), B_{z,w}^{(n)} = 1,\;\forall z,w \}$, where $\mathcal{E}(\mathbf{I}^{\text{q}}_{z})(\mathbf{x}_{z,n,w})$ is the keypoint representation extracted at $\mathbf{x}_{z,n,w}$ from the query image $\mathbf{I}^{\text{q}}_{z}$,  $\mathbf{B}^{(n)}\! \in\! \{0, 1\}^{Z\times W}$ denotes the selection indicator for the $n$-th type of predicted query keypoints, which is $\mathbf{B}^{(n)}\!=\!\argmax\nolimits_{\| \mathbf{B}'^{(n)} \|_{0}=\eta}\!\sum\nolimits_{z,w=1}^{Z\times W}\!p_{z,n,w}{B'}_{z,w}^{(n)}$.
% we solve
% \vspace{-0.2cm}
% \begin{equation}
% \mathbf{B}^{(n)}\!=\!\argmax\nolimits_{\| \mathbf{B}'^{(n)} \|_{0}=\eta}\!\sum\nolimits_{z,w=1}^{Z\times W}\!p_{z,n,w}{B'}_{z,w}^{(n)}.
% \end{equation}
Based on the candidate set $\mathcal{Q}^{(n)}$, we refine SKPs by the \emph{soft assignment} as
%% 1) Simple average
% \begin{equation}
%     (\mathbf{c}_{n})^*= \frac{1}{|\mathcal{S}^{(n)}| + |\mathcal{Q}^{(n)}|} \Big(\sum\nolimits_{\mathbf{\Phi} \in \mathcal{S}^{(n)}}\mathbf{\Phi} + \sum\nolimits_{\tilde{\mathbf{\Phi}} \in \mathcal{Q}^{(n)}} \tilde{\mathbf{\Phi}} \Big).
% \end{equation}
%% 2) Soft assignment
%\begin{equation}
  \begin{align}
   & \!\!\!\!\!
    \mathbf{c}_{n}^*\!=\!\frac{1}{\kappa|\mathcal{S}^{(n)}| \!+\!(1-\kappa)\sum\nolimits_{\tilde{\mathbf{\Phi}}\in \mathcal{Q}^{(n)}} p(\tilde{\mathbf{\Phi}}, \mathbf{c}_n)} \Big(\kappa\sum\nolimits_{\mathbf{\Phi} \in \mathcal{S}^{(n)}}\mathbf{\Phi}\nonumber\\
    &\qquad\qquad\qquad+(1-\kappa)\sum\nolimits_{\tilde{\mathbf{\Phi}} \in \mathcal{Q}^{(n)}} p(\tilde{\mathbf{\Phi}}, \mathbf{c}_n)\tilde{\mathbf{\Phi}} \Big),\!\!
  \end{align}
%\end{equation}
where $0\!\leq\!\kappa\!\leq\!1$ controls the amount of added pseudo-label information, and %$p(\tilde{\mathbf{\Phi}}, \mathbf{c}_n)=\text{softmax}(-\|\tilde{\mathbf{\Phi}}-\mathbf{c}_n\|/2\sigma^2)$,
$p(\tilde{\mathbf{\Phi}}, \mathbf{c}_n)\!=\!\exp(-\|\tilde{\mathbf{\Phi}}\!-\!\mathbf{c}_n\|/2\sigma^2)/\sum\nolimits_{i=1}^{N} \exp(-\|\tilde{\mathbf{\Phi}}\!-\!\mathbf{c}_i\|/2\sigma^2)$
is the assignment probability given $\mathbf{c}_n$. The updated SKP $\mathbf{c}_{n}^*$ is more representative and utilized to re-estimate query keypoints. % that tend to be more accurate. %After a number of iterations, the transductive FSKD could fully exploit the unlabeled data to enhance the performance.

\begin{table*}[!tb]
%\vspace{-0.3cm}
  \centering
  \caption{Results on few-shot keypoint detection for unseen species across four datasets. The harmonic mean of PCK scores on novel and base keypoint detection is reported. Symbol \textbf{+T}  refers to applying transductive inference during testing. \textbf{IT} is keypoint inference time per query image. The best and second best scores are marked in \bfa{red} and \bfb{blue}.
  %The results are the average of three runs over 1000 episodes.
  %PCK metric is used and the 95\% confidence intervals of all scores are below 1.2\%. The average of five subproblems in Animal pose dataset is reported. The results marked with $^*$ are copied from original paper. $^{**}$ and \textbf{+T} refer to apply finetuning or transductive inference during testing, respectively. The best and second best results are marked in \textbf{bold} and \underline{underline}.
  }
  \label{tab:one-five-shot-fskd-benchmark}
  \small
  % \resizebox{\textwidth}{40mm}{  % resize table
  % \begin{tabular*}{.95\linewidth}{@{\extracolsep{\fill}}c|l|cc|cc|cc|cc}
    \begin{tabular}{c|l|cc|cccccc|c|c|c}
      \multirow{2}*{\textbf{Shots}} & \multicolumn{1}{c|}{\multirow{2}*{\textbf{Model}}} & \multirow{2}*{\textbf{Params}} & \multirow{2}*{\textbf{IT}} & \multicolumn{6}{c|}{\textbf{Animal Pose Dataset}}  & \multirow{2}*{\textbf{CUB}} & \multirow{2}*{\textbf{NABird}} & \multirow{2}*{\textbf{AwA}} \\
                          &      &&& \textbf{Cat}  & \textbf{Dog}  & \textbf{Cow}  & \textbf{Horse} & \textbf{Sheep} & \textbf{Avg}  &      &      &      \\ \specialrule{.1em}{0em}{0em}%\hline%\midrule[1pt]
    \multirow{10}*{1-shot}&ProbIntr \cite{novotny2018self}&29.5M&3.8 ms&35.21 &30.01 &25.79 &24.87  &23.33  &27.84 &70.66 &57.63 &40.65 \\
                    &ProtoNet \cite{snell2017prototypical}&19.0M&3.4 ms&27.53 &23.01 &20.37 &18.03  &20.55  &21.90 &62.64 &48.84 &38.98 \\
                      &RelationNet \cite{sung2018learning}&21.1M&3.6 ms&30.89 &25.01 &22.29 &20.52  &22.91  &24.32 &66.08 &47.53 &30.64 \\
                  &WG (w/o Att.) \cite{gidaris2018dynamic}&19.0M&3.9 ms&30.70 &24.87 &23.58 &23.23  &22.63  &25.00 &63.64 &47.18 &38.85 \\
                             &WG \cite{gidaris2018dynamic}&23.2M&4.1 ms&31.02 &27.04 &24.25 &23.80  &23.40  &25.90 &65.65 &48.15 &39.45 \\
                               &FSKD-R \cite{Lu_2022_CVPR}&65.2M&3.9 ms&50.99 &45.29 &42.07 &43.13  &36.61  &43.62 &82.62 &66.89 &57.76 \\
                              & FSKD-D \cite{Lu_2022_CVPR}&65.2M&3.9 ms&54.30 &49.56 &46.06 &46.01  &39.94  &47.17 &82.51 &68.17 &65.56 \\
                              & \textbf{SalViT (DINO)} &50.9M&4.9 ms&55.75 &50.29 &46.90 &47.23  &41.66  &48.37 &84.50 &69.45 &\bfb{69.75} \\
                               &\textbf{SalViT (default)} &56.9M&4.5 ms&\bfb{57.00} &\bfb{50.94} &\bfb{48.13} &\bfb{47.87}  &\bfb{42.36}  &\bfb{49.26} &\bfb{84.62} &\bfb{69.58} &68.18 \\
                             &\textbf{SalViT (default)+T} &56.9M&6.6 ms&\bfa{59.94} &\bfa{54.82} &\bfa{52.06}  &\bfa{50.01} &\bfa{45.61}  &\bfa{52.49} &\bfa{88.36} &\bfa{72.47} &\bfa{70.79} \\ \hline\hline
    \multirow{10}*{5-shot}&ProbIntr \cite{novotny2018self}&29.5M&4.9 ms&43.83 &40.06 &33.41 &36.24  &29.47  &36.60 &78.78 &68.76 &53.23 \\
                    &ProtoNet \cite{snell2017prototypical}&19.0M&4.6 ms&35.23 &31.11 &30.25 &29.77  &28.42  &30.96 &71.81 &64.56 &46.43 \\
                      &RelationNet \cite{sung2018learning}&21.1M&4.6 ms&33.24 &28.75 &29.69 &26.38  &29.06  &29.42 &70.43 &55.58 &40.06 \\
                  &WG (w/o Att.) \cite{gidaris2018dynamic}&19.0M&5.0 ms&30.41 &28.13 &29.64 &32.15  &27.64  &29.59 &72.52 &60.70 &45.90 \\
                             &WG \cite{gidaris2018dynamic}&23.2M&5.1 ms&40.11 &36.14 &34.67 &35.61  &31.47  &35.60 &71.95 &60.98 &50.87 \\
                               &FSKD-R \cite{Lu_2022_CVPR}&65.2M&4.9 ms&59.43 &52.59 &53.13 &55.04  &46.43  &53.33 &86.99 &75.66 &71.59 \\
                              & FSKD-D \cite{Lu_2022_CVPR}&65.2M&4.9 ms&61.33 &55.62 &56.19 &55.11  &49.25  &55.50 &86.17 &75.50 &74.84 \\
                                 & \textbf{SalViT (DINO)} &50.9M&6.4 ms&62.03 &\bfb{58.65} &57.41 &56.69  &50.47  &57.05 &\bfb{88.18} &76.65 &\bfb{79.41} \\
                               &\textbf{SalViT (default)} &56.9M&5.5 ms&\bfb{62.24} &57.33 &\bfb{57.93} &\bfb{57.33}  &\bfb{50.50}  &\bfb{57.07} &87.76 &\bfb{77.08} &76.77 \\
                             &\textbf{SalViT (default)+T} &56.9M&7.6 ms&\bfa{64.49} &\bfa{59.29} &\bfa{60.39} &\bfa{59.91}  &\bfa{53.01}  &\bfa{59.42} &\bfa{89.81} &\bfa{78.85} &\bfa{81.03} \\
    \end{tabular}
%  }  % resizebox
\end{table*}

\section{Experiments}\label{sec:experiments}
\subsection{Settings}
\noindent\textbf{Datasets:} We conduct FSKD experiments on \emph{five} datasets. \emph{Animal pose} dataset \cite{cao2019cross} has five mammal species \emph{cat}, \emph{dog}, \emph{cow}, \emph{horse}, and \emph{sheep}, with 6000+ instances with keypoint annotations. \emph{CUB} \cite{WahCUB_200_2011} and \emph{NABird} \cite{van2015building} have 200 and 555 categories, respectively. \emph{AwA} \cite{banik2021novel} has 35 animal species with 10064 images. \emph{DeepFashion2} \cite{deepfashion2} is a large-scale clothing dataset with 13 categories and 8$\sim$39 landmarks.

%We conduct FSKD experiments on \emph{five} datasets. \emph{Animal pose} dataset \cite{cao2019cross} has five mammal species \emph{cat}, \emph{dog}, \emph{cow}, \emph{horse}, and \emph{sheep}, with 6000+ instances with keypoint annotations. \emph{CUB} \cite{WahCUB_200_2011} has 200 species, 15 keypoint annotations, and 11788 images. \emph{NABird} \cite{van2015building} is more complex and spans 555 categories, 11 keypoint annotations, and 48562 images. \emph{AwA} \cite{banik2021novel} has 35 animal species with 10064 images. \emph{DeepFashion2} \cite{deepfashion2} is a large-scale clothing dataset with 13 categories and 8$\sim$39 landmarks.

%\vspace{0.1cm}
\noindent\textbf{Metric:} The percentage of correct keypoints (PCK) \cite{yang2012articulated} is used. A predicted keypoint is correct if its distance to GT $d \leq \tau\!\cdot\! \max ( w_{\text{bbx}}, h_{\text{bbx}} )$, where $w_{\text{bbx}}$ and $h_{\text{bbx}}$ are the edges of object bounding box, and $\tau=0.1$.

%The percentage of correct keypoints (PCK) \cite{yang2012articulated} is used. A correct keypoint prediction requires its distance to groundtruth (GT) be less than $\tau\!\cdot\! \max ( w_{\text{bbx}}, h_{\text{bbx}} )$, where $w_{\text{bbx}}$ and $h_{\text{bbx}}$ are the edges of object bounding box. We set $\tau$  to $0.1$. %In addition, the harmonic mean of novel and base keypoint detection scores is also used.

%\vspace{0.1cm}
\noindent\textbf{Baseline Methods:} Firstly, we adapt \emph{ProbIntr}~\cite{novotny2018self} trained jointly by self-supervised keypoints and labeled query keypoints for comparison. Then, popular few-shot learning approaches \emph{ProtoNet}~\cite{snell2017prototypical}, \emph{RelationNet}~\cite{sung2018learning}, and \emph{WG} (with or w/o attention)~\cite{gidaris2018dynamic} are adapted to FSKD. Moreover, \emph{FSKD-R}~\cite{Lu_2022_CVPR} and \emph{FSKD-D}~\cite{Lu_2022_CVPR} are compared, which respectively uses random or default auxiliary keypoints during training. Following \cite{Lu_2022_CVPR}, we also use default auxiliary keypoints to boost keypoint generalization. We present two models for comparison, which are \emph{SalViT (default)} and \emph{SalViT (DINO)}. All is same except the former exploits CNN as backbone and MNL saliency \cite{zhang2018deep}, while the latter uses pre-trained DINO \cite{caron2021emerging} and its class token attentiveness as saliency.

\comment{
Firstly, we adapt the keypoint detection approach that presents probabilistic introspection matching loss (\emph{ProbIntr}) \cite{novotny2018self} for FSKD comparison. Since original ProbIntr is self-supervised detector, for fairness, we train ProbIntr jointly by self-supervised keypoints and labeled query keypoints. Then, the popular few-shot learning approaches \emph{ProtoNet} \cite{snell2017prototypical}, \emph{RelationNet} \cite{sung2018learning}, \emph{WG} (with or w/o attention) \cite{gidaris2018dynamic}, and Two-stage Finetuning Approach (\emph{TFA}) \cite{wang2020frustratingly} are adapted to FSKD. Moreover, two competitive models proposed by \cite{Lu_2022_CVPR} are compared, \ie \emph{FSKD (rand)} and \emph{FSKD (default)}, which have same architecture but differed in adopting two kinds of auxiliary keypoints into training to boost keypoint generalization, \ie interpolating three auxiliary keypoints per path and randomly sampling six paths constructed by arbitrary body parts, or select six default limbs as paths. For comprehensive comparisons, we also build two variants \emph{SalViT (rand)} and \emph{SalViT (default)} for few-shot keypoint detection.
}

\noindent\textbf{Implementation Details:}
% introduce some details and common hyper-parameters for SalViT
The input size to all models is $384 \times 384$. For SalViT (default) model, the backbone uses ResNet50 \cite{he2016deep}, which has 23M params and is commonly used in detection with transformer, \eg, DETR \cite{carion2020end}. For SalViT (DINO), we choose DINO ViT-S/16 (21M params) pre-trained on ImageNet 1K \cite{deng2009imagenet} as backbone. The first 10 layers of DINO ViT are frozen and the class token attentiveness is taken from frozen layers due to better generalization that produces better ``saliency'' (see Sec.~\ref{app:saliency} in \textbf{Suppl.} for vis.) By default, we use one SalViT with RBF-based M-SA, set the temperature $\beta\!=\!1$ and the degree of masking $J\!=\!1$. To keep SalViT light-weight, token dim. is set to 384 as in ViT-S. All models are trained with 40k episodes. The scores are averaged over 1000 episodes in testing. In transductive inference, the number of test query images is 60.

\begin{table}[!tb]
  \centering
  \caption{1-shot/5-shot novel keypoint detection on DeepFashion2. %\cite{deepfashion2}. %The upper-body clothing categories (6 categories) are set as seen categories and lower-body clothing categories (7 categories) are set as unseen categories.
  The metric of NE (normalized error $d'=d / \max(w_{\text{img}}, h_{\text{img}})$) and PCK %($\tau\cdot\max(w_{\text{bbx}}, h_{\text{bbx}})$, $\tau=0.1$)
  with 95\% confidence intervals are used. %MetaCloth \cite{ge2021metacloth} tests on 700 episodes and uses 12 gradient steps of finetuning. Our model tests on 1000 episodes with 10 gradient steps of fine-tuning.
  The results marked with $^*$ are from \cite{ge2021metacloth}. `-': no reported results.
  }
  \label{tab:fskd-deepfashion2}
  % \small
  \resizebox{.48\textwidth}{!}{  % resize table 0.8-14/1.0-18
  \begin{tabular}{l|cc|cc}
    % \toprule[1pt]
    % \hline
    \multicolumn{1}{c|}{\multirow{2}*{Method}}  & \multicolumn{2}{c|}{1-shot}          & \multicolumn{2}{c}{5-shot}  \\
                          & PCK ($\uparrow$) & NE ($\downarrow$) & PCK ($\uparrow$) & NE ($\downarrow$) \\ \specialrule{.1em}{0em}{0em}
    MAML$^*$~\cite{finn2017model}&  -               & 0.215 $\pm$ 0.023 &  -               & 0.126 $\pm$ 0.017 \\
    ProtoNet$^*$~\cite{snell2017prototypical}&  -   & 0.179 $\pm$ 0.042 &  -               & 0.135 $\pm$ 0.039 \\
    WG$^*$~\cite{gidaris2018dynamic}&  -            & 0.185 $\pm$ 0.051 &  -               & 0.160 $\pm$ 0.047 \\
    FSKD~\cite{Lu_2022_CVPR}     & 33.04 $\pm$ 1.80 & 0.149 $\pm$ 0.005 & 44.92 $\pm$ 1.72 & 0.111 $\pm$ 0.004 \\
    MetaCloth$^*$~\cite{ge2021metacloth}&  -        & 0.145 $\pm$ 0.021 &  -               & 0.096 $\pm$ 0.015 \\
    \textbf{SalViT (DINO)}     & 35.59 $\pm$ 1.38 & 0.134 $\pm$ 0.004 & 48.00 $\pm$ 1.31 & 0.089 $\pm$ 0.002 \\
    \textbf{SalViT (default)}  & \bf{36.15 $\pm$ 1.46} & \bf{0.130 $\pm$ 0.004} & \bf{50.52 $\pm$ 1.32} & \bf{0.087 $\pm$ 0.003} \\
    % \textbf{SalViT (default)} & \bfa{35.01 $\pm$ 1.47} &\bfa{0.130 $\pm$ 0.004} & \bfa{50.87 $\pm$ 1.18} & \bfa{0.093 $\pm$ 0.002} \\
    % \bottomrule[1pt]
    %\hline
  \end{tabular}
 }  % resizebox
  % \vspace{-0.3cm}
\end{table}
\begin{figure*}[!tb]
\vspace{-0.3cm}
  \centering
  \includegraphics[width=.95\linewidth]{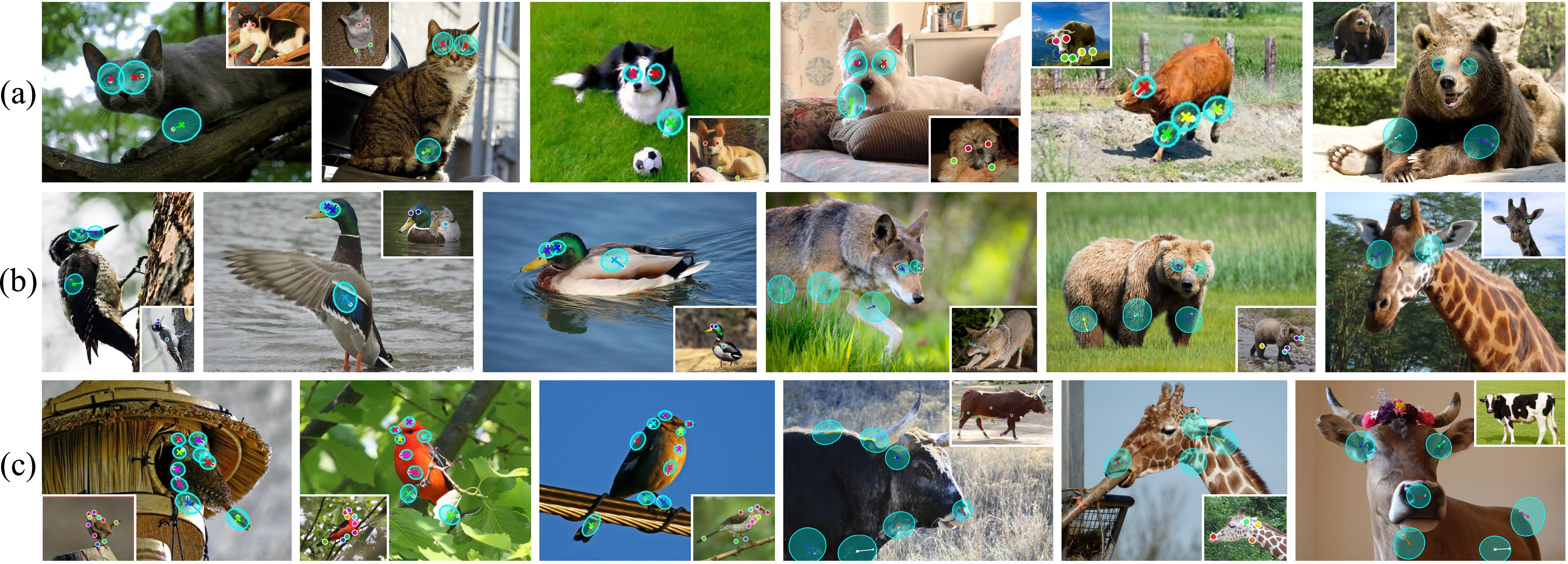}
  \caption{Visualization of few-shot keypoint detection for unseen species. Rows (a) and (b) show the results on detecting novel keypoints while (c) shows detected base keypoints. The support image \& keypoints are in the corner of each query image. Dots, tilted crosses, and ellipses refer to GT, predicted keypoints and uncertainty, respectively. The line that links prediction and GT reflects the localization error. %The small image on the corner is the support while the big image is the query. The support keypoints and GT query keypoints are shown by circle dots. Our FSKD keypoint prediction is shown by tilted cross centered by an ellipse which represents the localization uncertainty with 99.7\% confidence. Not all support keypoints have the corresponding keypoints in query image as the GT may not exist. The line segment that connects from prediction to GT reveals the localization error. The rows (a)$\sim$(b) show the results on detecting novel keypoints while row (c) shows results for base keypoints.
%The rows (a)$\sim$(c) show the results of detecting novel keypoints, where rows (a)$\sim$(b) are for each species in animal pose dataset while (c) is for CUB and NABird which occupies three images respectively. The row (d) shows the detection results for base keypoints. The goal of uncertainty is to bound the GT keypoints.
}
  \label{fig:fskd-visual}
  % \vspace{-2pt}
\end{figure*}

\subsection{Comparisons on Few-shot Keypoint Detection}
% setting for species split & keypoint split
% quantitative results (maybe we can split the results based on inductive or transductive inferences)
% visualize FSKD (qualitative results)

% To set up FSKD experiments, the species and annotated keypoints are split into \emph{disjoint} seen and unseen species, base and novel keypoints. Then the FSKD models are trained on seen species with base keypoints, while testing on unseen species with novel keypoints. %, which is a quite challenging due to large domain shift on keypoints and species. %This setting is very challenging due to the large domain shift on keypoints and species and is able to validate the generalization ability of models.

All compared FSKD models are trained on seen species with base keypoints, and tested on unseen species with novel (or base) keypoints. The split settings of seen and unseen species, novel and base keypoints across five datasets are same to work \cite{Lu_2022_CVPR}. For Animal pose dataset, one animal species is alternately chosen as unseen species for testing while the remaining four as seen species for training, which yields five subproblems. For AwA, 25 species are for training and 10 for testing. For DeepFashion2, 6 upper-body clothing categories are set as seen categories and 7 lower-body clothing categories are set as unseen ones \cite{ge2021metacloth,Lu_2022_CVPR}.

Table~\ref{tab:one-five-shot-fskd-benchmark} shows our model variants SalViT (default) and SalViT (DINO) consistently obtain higher harmonic scores of novel and base keypoint detection across all tasks, and enjoy significant improvements. For 1-shot setting, our SalViT (default) outperforms previous state-of-the-art model FSKD-D \cite{Lu_2022_CVPR} by 2.09\%  in Animal Avg, 2.11\% in CUB, 1.41\% in NABird, and 2.62\% in AwA. Our SalViT (DINO) performs very well on AwA, and achieves 4.19\% improvement in 1-shot setting and 4.57\% in 5-shot setting. Moreover, both model variants have fewer parameters than FSKD-D at the expense of slightly longer inference time. The higher scores explain the effectiveness of SalViT in detecting few-shot keypoints by learning foreground patch relations. Even though the saliency prior of DINO is not as good as MNL prior (see Table~\ref{tab:ious}), it achieves very competitive scores. We conjecture it may be due to the function of morphology learner (ML) inside SalViT, which refines lower quality saliency priors.

By applying transductive inference, the performance is further improved. %For example,
Compared to our SalViT (default), the transductive version obtains further gains of 3.23\% in Animal Avg, and 3.74\% in CUB in 1-shot setting. The stronger performance indicates that the unlabeled data enhances keypoint representations and benefits FSKD. Table~\ref{tab:fskd-deepfashion2} shows our SalViT achieves higher PCK scores yet lower NE error in DeepFashion2 dataset. Fig.~\ref{fig:fskd-visual} shows the qualitative results of our approach, which manifests its effectiveness to localize keypoints and estimate keypoint uncertainty.

\begin{table}[!t]
  \centering
  \caption{Ablation study of SalViT in 1-shot novel keypoint detection. The result on Animal is the average of five subproblems.
  %Ablation study of SalViT in 1-shot novel keypoint detection. M-SA: masked self-attention; PE: position encoding; APE$^*$: learnable absolute PE \cite{dosovitskiy2020image}; RPE$^*$: relative PE same to SwinT \cite{liu2021swin} but different to ours. The result on Animal dataset is the average of five subproblems. %All results are obtained by average of three runs and each run is the average of 1000 episodes (95\% confidence intervals are all below 1.2\%).
  %Ablation study of SalViT (in the setting of 1-shot novel keypoint detection). ML: Morphology learner (w/o means no using ML and pre-IP); PE: Position encoding; RPE: relative position encoding; APE: absolute position encoding (we use learnable one, namely APE1). All results are obtained by average of three runs and each run is the average of 1000 episodes (95\% confidence intervals are all below 1.2\%).
  }
  \label{tab:ablation-study}
  \small
  % \resizebox{\hsize}{20mm}{  % resize table
  \begin{tabular}{l|cccc}
    %\hline
    Models                 &Animal   & CUB  & NABird& AwA  \\ \specialrule{0.1em}{0em}{0em}
    \emph{DINO w. ViT}     &44.03    &76.96 &56.11  &66.25 \\
    \emph{DINO w. SalViT}  &\textbf{45.71}    &\textbf{80.08} &\textbf{57.19}  &\textbf{68.71} \\ \specialrule{0.1em}{0em}{0em}
    %\emph{w/o ML}          &43.72    &78.63 &56.27  &66.89 \\ \hline\hline
    \emph{CNN-only}        &44.25    &76.54 &54.71  &61.75 \\
    \emph{ViT-only}        &43.19    &76.71 &54.44  &62.53 \\
    \emph{CNN+ViT}         &44.34    &76.62 &54.57  &63.22 \\
    % \emph{Non-local}       &         &      &       &      \\
    % \emph{Dynamic Conv.}   &         &      &       &      \\
    \emph{SalViT}          &\textbf{46.61}    &\textbf{79.52} &\textbf{57.36}  &\textbf{67.74} \\ \specialrule{0.1em}{0em}{0em}
    \emph{SalViT (w/o ML)} &44.79 &77.70 &56.66  &64.19 \\
    \emph{SalViT (w/o PE)} &45.02 &77.03 &55.41  &65.23 \\
    % \emph{SalViT (w. APE)} &44.34    &76.82 &54.82  &62.73 \\
    % \emph{SalViT (w. RPE$^*$)} &45.81    &78.41 &55.97  &65.11 \\
    %\hline
  \end{tabular}
  % }  % resizebox
  \vspace{-0.3cm}
\end{table}

\subsection{Ablation Studies}\label{sec:ablation_study}
% 1) ablation. CNN only, vanilla vit vs. salvit
% 2) influence of salvit dim vs. CNN dim (we can use ratio to represent their proportion)
% 3) influence of salvit num
% 4) influence of PN value. PN learned vs. PN manual (PN Value=1.0, 0.7, etc.)
% 5) influence of saliency detectors (draw a graph show 1/5-shot SalViT using diff. sal. det., see PPT)
% 6) comparison of different transductive FSKD ( A) score+simple avg; B) score+soft assignment; C) [0.5score+0.5simi]+simple avg; D) [0.5score+0.5simi]+soft assignment)
% visualize attention maps, morphology changed saliency maps.
% (draw graphs, opt.) compute statistical PN value per layer (over 1000 episodes, on different datasets)

%The effectiveness of each component is validated using
The ablations are performed using our SalViT (default) model on novel keypoint detection for unseen species.

\noindent\textbf{Salient Patch Relations:} When using DINO as backbone, Table~\ref{tab:ablation-study} (top part) shows SalViT (DINO) greatly beats DINO with vanilla ViT. In CUB, SalViT (DINO) enjoys 3.12\% gain. When using CNN as backbone,  compared to CNN-only (relies on neighborhood information), or ViT-only (relies on global patch relations), Table~\ref{tab:ablation-study} (middle part) shows our SalViT consistently performs best, \eg, 4.52\% improvements in AwA than CNN+ViT, validating the effectiveness of capturing salient patch relations by our soft M-SA.

\noindent\textbf{Morphology Learner (ML):} When turning off ML in SalViT, we observe losses across four datasets (lower part, Table~\ref{tab:ablation-study}). The similar losses appear if position encoding (PE) is not used. These observations reflect the importance of adapting receptive field in M-SA and incorporating spatial information for keypoint inference. Moreover, we manually set or automatically learn PN value in ML. Table~\ref{tab:pn-value-influence} shows that morphology learning  improves scores. % and ML could adapt accordingly.

\noindent\textbf{Transductive Inference:} As transductive inference relies on the quality of pseudo labels, we firstly capture the upper bound of performance using the GT keypoints of queries to select high-quality pseudo labels. Then, we compare the employed soft assignment (SM) against the simple average-based SKP refinement. Table~\ref{tab:transductive} shows our approach yields higher scores closer to the upper bound.

\begin{figure}[!t]
\vspace{-0.3cm}
  \centering
  \includegraphics[width=\linewidth]{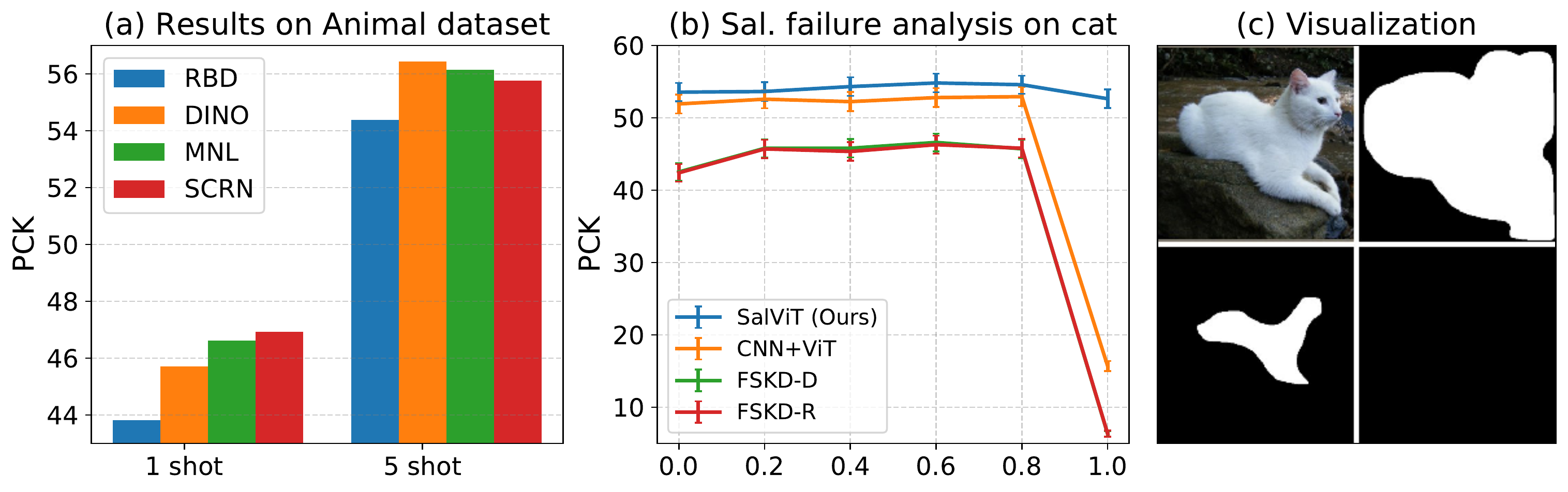}
  \caption{Results on SalViT with (a) different saliency detectors \& (b)-(c) simulated saliency failures by varying saliency thresholds.
  % diff. sal. detectors: 1/5 shot novel keypoint detection;
  % sal. failure: 1-shot novel keypoint detection on cat subproblem.
  }\label{fig:saliency-detectors-failures}
\end{figure}

\noindent\textbf{Saliency Detectors \& Saliency Failures:} % Different saliency detectors & saliency failure comes here
% in addition to the deep unsupervised saliency detector MNL, we also investigate RBD which performs best among traditional sal. detectors \cite{borji2015salient}, and the deep supervised SCRN. A glimpse of visual effects can be found in Suppl. Material.
Fig.~\ref{fig:saliency-detectors-failures}(a) shows SalViT with unsupervised MNL or DINO yields  performance close to SCRN in 1-shot setting (and better in 5-shot). Although better saliency prior improves performance, our ML leverages saliency maps efficiently, mitigating the influence of low-quality maps.
Moreover, we test the trained model under simulated saliency failures. Fig.~\ref{fig:saliency-detectors-failures}(b) shows performance of SalViT drops a little under large saliency failures. If foreground (FG) is available, FG relations between tokens are captured by our M-SA. When FG is removed, then each token uses self-relation (identity matrix in Eq.~\ref{eq:attention-mask}). Such a property of M-SA ensures signal flows through attention even for bad sailency failures. %, as FG information is not completely lost, only FG relations are not leveraged.
We also equip other models with saliency by re-weighting patch features for refinement but that helps little under failures. %. They slightly boost scores when properly thresholding but fails under bad saliency.

% in Eq.~\ref{eq:attention-mask} \& Eq.~\ref{eq:soft-masking}
\comment{
Firstly, we investigate the effects using unsupervised saliency detectors (MNL, DINO, and RBD) and the supervised SCRN. Fig.~\ref{fig:saliency-detectors-failures}(a) shows SalViT with MNL or DINO yields closer performance to SCRN in 1-shot setting and better on 5-shot. Intuitively, better saliency priors yield better performance. However, our ML can leverage saliency maps efficiently to mitigate the influence of low-quality maps. Moreover, we test the trained model under simulated saliency failures. Fig.~\ref{fig:saliency-detectors-failures}(b) shows our SalViT drops only tiny bit under large saliency failures. Even when saliency map is reversed and foreground (FG) mask is destroyed. Our M-SA (attention) only loses the long-range relations on FG patches. The way our attention in \textbf{Eq.~(5)(6)} works is that if FG is available, FG relations between tokens are captured. When FG is removed, then each token's self-relation is strenghtened (see \textbf{identity} matrix in Eq. (5)). Such a \textbf{property of M-SA } ensures signal flow through attention even for such a bad failures as FG information is not completely lost, only FG relations are not leveraged. We also equip other models with saliency by re-weighting CNN/patch features for refinement. They slightly boost scores when properly thresholding but fails under bad saliency.
}

\begin{table}[!t]
\vspace{-0.3cm}
  \centering
  \caption{Influence of power normalization (PN) value on SalViT in FSKD. The \emph{auto} means using learnable PN, namely turning on the morphology learner (ML) in SalViT. %The saliency map is pre-processed by image processing to ensure smoothness for PN. %The best and second best results are marked in \textbf{bold} and \underline{underline}.
  %Influence of power normalization (PN) value to SalViT in FSKD. The \emph{auto} means the results from learned PN, namely using the ML (Morphology Learner). The saliency map all are pre-processed by image processing (pre-IP) for sake of PN. The best and second best results are marked in \textbf{bold} and \underline{underline}.
  }
  \label{tab:pn-value-influence}
  \small
  % \resizebox{\hsize}{6mm}{  % resize table
  {
  \setlength{\tabcolsep}{3pt}
  \begin{tabular}{c|cccccc|c}
    % \toprule[1pt]
    %\hline
    Datasets&0.2  &0.5  &0.7  &0.8  & 1.0 & 2.0 &auto \\ \specialrule{0.1em}{0em}{0em}%\midrule[1pt] %\hline
    Animal  &43.56&44.14&45.32&44.63&44.81&44.16&\textbf{46.61}\\ %\midrule[0.5pt] %\hline
    AwA     &61.53&61.57&63.58&62.94&62.23&61.29&\textbf{67.74}\\
    % \bottomrule[1pt]
    %\hline
  \end{tabular}
  }
% }  % resizebox
\end{table}

\begin{table}[!t]
  \centering
  \caption{Ablation of prototype refinement approaches in transductive FSKD. Harmonic scores of 1-shot setting are reported. %We use the harmonic results ($2ab/(a+b)$).
  }
  \label{tab:transductive}
  \resizebox{0.4\textwidth}{!}{  % resize table
  \begin{tabular}{l|ll}
  %\hline
  Method                      &Animal        & AwA  \\ \specialrule{0.1em}{0em}{0em}
  \emph{Inductive} (baseline) &49.26         &68.18 \\
  \emph{Trans.+Avg}           &50.27 (+1.01) &69.17 (+0.99) \\
  \emph{Trans.+SM} (ours)     &\textbf{52.49 (+3.23)} &\textbf{70.79 (+2.61)} \\
  \emph{Trans.+GT}            &54.23 (+4.97) &73.09 (+4.91) \\ % upper bound
  %\hline
  \end{tabular}
  }
\end{table}

\subsection{FSKD Under Occlusions}
% construct an experiment to analyse the performance of salvit under occlusion
% compared models: 1) CNN feature only model; 2) Vanilla ViT-only feature model; 3) CNN+ViT feature model; 4) SalViT feature model
% interpolation: default.
% occlusion rate: r = 0%, 10%, 30%, 50%, 70%, 100%
% datasets: animal pose, CUB, NABird, etc.
\begin{table}[!tb]
 %\vspace{-0.3cm}
  \centering
  \caption{Ablations of occlusion handling with various masking strategies. %Results are tested under occlusion of \textbf{background crop}. while the models were trained using masks of \emph{gray box}.
  Tick \Checkmark indicates applying foreground patch masking over RGB image, saliency map, or CNN patch features; no tick means no masking. %\XSolidBrush means no masking.
  Each row indicates a masking strategy for FSKD model training. The model is tested under various occlusion probabilities $p$ %($p=0\%, 10\%, 30\%, 50\%, 70\%, 100\%$)
  of query keypoints. The higher the $p$ is, the more severe the occlusions are. Average PCKs of novel keypoint detection on \emph{five} subproblems in Animal pose dataset are given.
  }
  \label{tab:occlusion-bgbox}
  % \small
  % \resizebox{\linewidth}{15mm}{
  \begin{tabular}{ccc|cccccc}
    % \toprule[1pt]
    % \hline
    RGB         & Sal.       & Feat. &0\%  &30\% &50\% &100\% \\ \specialrule{0.1em}{0em}{0em} %\midrule[1pt]%\hline
                &            &       &45.60&38.49&34.05&22.82 \\ %\hline
    \Checkmark  &            &       &45.60&40.49&37.94&30.27 \\
    \Checkmark  &\Checkmark  &       &46.28&40.68&37.70&29.77 \\
    \Checkmark  &\Checkmark  &\Checkmark &46.04&40.75&37.42&29.13 \\
              % &\Checkmark  &           &45.43&38.48&34.30&22.63 \\
                &\Checkmark  &\Checkmark &46.18&38.84&34.63&23.74 \\
    % \bottomrule[1pt]
    %\hline
  \end{tabular}
  % } % resizebox
  % \vspace{-0.2cm}
  \vspace{-0.3cm}
\end{table}
\begin{table}[!tb]
\vspace{-0.3cm}
  \centering
  \caption{Ablations of various masking and alignment (MAA) strategies. The results are the average of five subproblems in Animal pose dataset.  Non-Occl means adding the prediction loss from non-occluded images. Y: applied; N: not applied.  % tested under occlusion of \emph{background crop}.
  }
  \label{tab:MAA-bgcrop}
  \begin{tabular}{cl|ccc}
    %\hline % average
    Mask & Alignment        &0\%  &50\% &100\% \\ \specialrule{.1em}{0em}{0em}%\hline
      N  & N               &45.60&34.05&22.82 \\ %\hline
      Y  & N               &46.28&37.70&29.77 \\ %\hline
      % Y  & MAE              &     &     &      \\
      Y  & SimMIM \cite{xie2022simmim}&46.10&38.15&30.61 \\
      Y  & Non-Occl Loss          &45.65&37.15& 28.22 \\
      Y  & Feat. Align (MMD)      &45.44&36.68& 28.52 \\
      Y  & Feat. Align ($\ell_2$) &46.29&38.21& 29.68 \\
      Y  & Feat. Align ($\ell_1$) &\textbf{46.80}&38.22& 30.13 \\
      Y  & Prob. Align      &46.74&\textbf{38.40}& \textbf{32.04} \\
      % Y  & Feat. (MSE)+Prob.&45.71&39.03& 31.72 \\
    %\hline
  \end{tabular}
  \vspace{-0.3cm}
\end{table}

%ViT is reported to have strong ability to handle occlusions and noise thanks to self-attention. However, such ability would normally be still weak if no occlusion data used during training. On the other hand, it remains unclear whether our masked self-attention (M-SA) is effective to learn occlusion handling ability in FSKD.
Below we investigate if our masked self-attention is robust to keypoint occlusions, %To answer this question, we test model robustness to occlusions and investigate
and consider various strategies to improve the  ability of our model to handle occlusions.

% Vision transformer is reported in literature that has strong ability to handle occlusions and noise due to the advantages of self-attention. However, such ability would normally be weak if no occlusion data specifically used for training. On the other hand, it remains unclear whether our masked self-attention (M-SA) is effective to learn occlusion handling ability in FSKD. To answer this question, we test model robustness towards occlusions and investigate the strategies to improve the occlusion handling ability of proposed models.

% need to revise (masked SA-->mask occlusions/occlusion patch is surpressed (analysis)-->RPE function)(turn off morphology learner)
% analyze each masking strategy's meaning
Firstly, we test the normally-trained FSKD model under different occlusion probability to query keypoints. %The occlusion types include  \emph{gray box}, \emph{average-pixel box} (a box filled by average pixel), or \emph{background crop}.
Each testing occlusion block is centered at query keypoint and has random size with an area ratio to object bounding box between $0.01$ and $0.04$ and aspect ratio between $0.7$ and $1.4$. Secondly, we explore  %endow the model with occlusion handling ability
masking different cues during training: RGB image, saliency map, and CNN patch features. Between $2$ and $12$  foreground patches in RGB image are randomly masked with gray box, saliency and CNN features are masked with zero values.
%For instance, randomly masking the foreground patches in RGB image with gray box, masking saliency and CNN feature with zeros.
The FSKD model is trained by various masking strategies and tested under occluded query keypoints. %Table~\ref{tab:occlusion-bgbox} shows the PCK scores of different models under different keypoint occlusion probability.
Table~\ref{tab:occlusion-bgbox} shows that with no masking applied, the normally-trained FSKD model is unable to handle occlusions well. Masking RGB image, RGB + Saliency, RGB + Saliency + CNN patch features show higher yet similar results towards occlusions of \emph{background crop} even though trained with occlusions of \emph{gray box}. % which indicates they have learned the occlusion handling ability.
%More test results on \emph{gray box} and \emph{average-pixel box} occlusions are in Section \ref{app:more-results-of-occlusions}. %Ideally a good occlusion handling ability should handle various occlusions.
%Moreover, when RGB + Saliency are masked, our M-SA is still effective to learn such occlusion ability. It is worthy to note that RGB image requires to be masked to acquire better occlusion ability.
Moreover, on top of masking inputs, we investigate different alignment approaches: 1) alignment in \emph{image space} using SimMIM~\cite{xie2022simmim}; 2) alignment in \emph{predictions} by adding the loss over non-occluded image (\ie Non-Occl); 3) alignment in \emph{feature space} with various distance metrics including $\ell_1$, $\ell_2$, and MMD~\cite{borgwardt2006integrating}; and 4) alignment in \emph{probability maps}. Table \ref{tab:MAA-bgcrop}  shows that using the probability alignment is the most effective, boosting scores by up to 2\% compared to masking-only, and by $10\%$ compared to the normally trained model with no MAA strategy under 100\% occlusions.
% observations: Prob. alignment is most effective and feat align is second most effective. Choosing the distance metric when performing feat. alignment is important. We observe l1/l2 is better than MMD.

%The occlusion handling ability of FSKD models trained by different masking strategies. In training, the masking operation is applied (to RGB image, saliency map, or CNN patch feature) with a probability of 0.5. When performing masking, there are $2\sim 12$ foreground patches are randomly masked (RGB image is masked with gray patch, saliency and CNN feature is masked with 0). During testing, we randomly occlude each keypoint with a probability $p=0\%, 10\%, 30\%, 50\%, 70\%, 100\%$, where the occlusion block has random size, with area ratio to object bounding box $0.01\sim 0.04$, aspect ratio $0.7 \sim 1.4$. In this table, occlusion block uses \textbf{background crop}.

\section{Conclusions}\label{sec:conclusion}
We have introduced a saliency-guided vision transformer, SalViT, for few-shot keypoint detection. The key insight is to learn the foreground patch relations via SalViT and use such relations for similarity learning in FSKD. The SalViT includes a uniquely designed masked self-attention (M-SA) and a morphology learner (ML), which together constrain self-attention within salient regions while allowing adaptation. Moreover, we have presented a transductive FSKD variant in addition to the standard inductive inference. The FSKD benchmarks across five datasets highlight that our FSKD with SalViT achieves competitive results. The extensive experiments show the effectiveness of M-SA and ML. The occlusion setting reveals that our SalViT-based model with proper masking and alignment handles occlusions of keypoints well. Finally, saliency detectors and backbone can be readily replaced by DINO.

{\small
\bibliographystyle{ieee_fullname}
\bibliography{refs}
}

%==========================================================================
%used to build the title for supplementary materials
%added by Peter
\newpage
\appendix
% \title{Saliency-guided Vision Transformer for Few-shot Keypoint Detection (Supplementary Material)}
\title{From Saliency to DINO: Saliency-guided Vision Transformer for Few-shot Keypoint Detection (Supplementary Material)}
%used to build the title for supplementary materials
%==========================================================================
\author{Changsheng Lu$^{1}$, Hao Zhu$^{1}$, Piotr Koniusz$^{2,1}$\\
  $^{1}$The Australian National University \quad
   $^2$Data61/CSIRO\\
%College of Engineering and Computer Science, Australian National University\\
{\tt\small \{ChangshengLuu,allenhaozhu\}@gmail.com, Piotr.Koniusz@data61.csiro.au}
%
% \author{Changsheng Lu$^{\dagger}$, Hao Zhu, Piotr Koniusz\textsuperscript{\textasteriskcentered}$^{,\S, \dagger}$\\
%   $^{\dagger}$The Australian National University \quad
%    $^\S$Data61/CSIRO\\
% %College of Engineering and Computer Science, Australian National University\\
% %Institution1 address\\
% {\tt\small ChangshengLuu@gmail.com, firstname.lastname@anu.edu.au}
%\and
}
\maketitle

% \textbf{\color{teal} We will release the source code to aid the future research and re-producibility by the vision community.}

\section{Further Details for Approach}
% \subsection{Notations}\label{app:notations}
% Capitalised bold symbols denote matrix, lowercase bold symbols denote vectors, and regular fonts denote scalars. We define $\mathbf{1}$ as `all ones' matrix or vector, $\mathbb{I}$ as identity matrix, and $A_{i,j}$ as the $(i,j)$-th entry of $\mathbf{A}$. Operator ``$:$'' extracts $i$-th vector $\mathbf{X}_{i,:}$ from matrix $\mathbf{X}$. $\text{Diag}(\cdot)$ embeds a vector to form a diagonal matrix. Symbol $\|\cdot\|_0$ means $l_0$ norm, representing the summation of entries.

%$\text{Diag}^{\dagger }(\cdot)$ puts the matrix diagonal into a vector.

\subsection{Visualizations of Various Saliency Priors}\label{app:saliency}
\begin{figure*}[!tb]
  \centering
  \includegraphics[width=\linewidth]{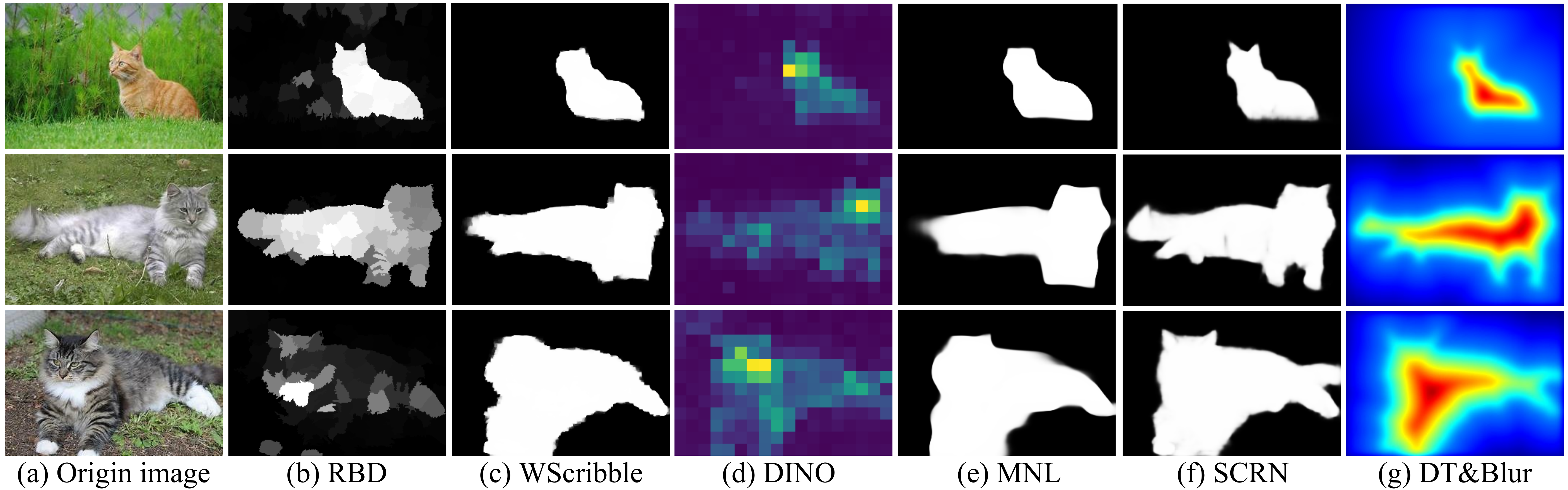}
  \caption{Examples of saliency prior masks generated by different models. The first column shows a selection of input images. Columns (b) to (f) show masks obtained with RBD \cite{zhu2014saliency}, WScribble \cite{zhang2020weakly}, DINO \cite{caron2021emerging}, MNL \cite{zhang2018deep}, and SCRN \cite{wu2019stacked}. The rightmost column shows the pre-processed saliency maps of coloumn (f) after the distance transform (DT) and blur. The colorized maps in the last column are for better visualization of saliency diffusion from foreground to background. %, which is useful for morphology learning.
  %originated from SCRN (\ie image (d)), where the colorized version is adjunctive for better visualization of the saliency diffusion from foreground to background.
  }
  \label{fig:saliency-map-samples}
\end{figure*}
Fig.~\ref{fig:saliency-map-samples} shows  examples of saliency maps produced by unsupervised (RBD \cite{zhu2014saliency}, DINO \cite{caron2021emerging}, and MNL \cite{zhang2018deep}), weakly-supervised (WScribble \cite{zhang2020weakly}), and supervised detectors (SCRN \cite{wu2019stacked}). As one can see, saliency maps usually capture foreground objects. When the foreground and background have distinct colors and textures, both traditional and deep learning based detectors separate the salient objects well from backgrounds (top two rows, Fig.~\ref{fig:saliency-map-samples}). If foreground and background are similar in color/texture, RBD may fail (third row, Fig.~\ref{fig:saliency-map-samples}). We notice that MNL and DINO perform well despite they are entirely trained without the human-annotated labels. %using the labels from multiple unsupervised detectors rather than human-annotated masks.

\subsection{Analysis of Saliency Detectors}
% Parameters and Inference time of Saliency Detectors
Table~\ref{tab:params-sal-detectors} shows MNL enjoys very fast inference speed at 1.67 ms/img with 16.3M parameters. DINO is the fastest (0.03 ms/img) and has no extra parameters as its saliency is produced from the DINO backbone itself. Compared to other models listed in the table, MNL and DINO are very efficient in terms of the inference time (MNL and DINO are very fast) and the low number of parameters (generating saliency priors with DINO does not introduce extra parameters at all when DINO is used as the backbone). In our implementation, the saliency is pre-extracted once to facilitate  easy training.

%          params, inference time for each Saliency Detectors
% RBD      : 0,      0.26539  s (3.768   img/s)
% WScribble: 15.5M,  0.029598 s (33.786  img/s)
% DINO     : 0,      0.000025 s (40000   img/s)
% MNL      : 16.3M,  0.001671 s (598.444 img/s)
% SCRN     : 25.2M,  0.019246 s (51.9588 img/s)
\begin{table}[!b]
  \centering
  \caption{Additional parameters and inference time (IT) of various saliency detectors.}
  \label{tab:params-sal-detectors}
  \setlength{\tabcolsep}{2.2pt}
  \begin{tabular}{c|ccccc}
      Sal. Det.         & RBD    & WScribble & DINO   & MNL   & SCRN  \\ \specialrule{.1em}{0em}{0em}
      Add. Params (M)   & 0      &15.5       &0       &16.3   &25.2   \\
      Add. IT (ms)      &265.39  &29.60      &0.03    &1.67   &19.25 \\
  \end{tabular}
\end{table}

\subsection{Architecture Details}\label{app:architecture-details}
Let $(k, s, p)$ be the parameters of a convolution (conv.), where $k$ is kernel size, $s$ is stride, and $p$ is padding.

\vspace{0.1cm}
\noindent\textbf{Saliency Embedding Module (SEM)} includes a large-stride conv., three small-stride conv., and a $1\times 1$ conv. Except the last $1\times 1$ conv., each layer is followed by a ReLU to introduce non-linearity.

\comment{
\noindent\textbf{Saliency Embedding Module (SEM)} includes a large-stride conv. with $\{k=32, s=32, p=0\}$, three small-stride conv. with $\{k=3, s=1, p=1\}$, and a $1\times 1$ conv. Except the last $1\times 1$ conv., each layer is followed by a ReLU to provide non-linearity.
}

\vspace{0.1cm}
\noindent\textbf{SalViT based Feature Encoder $\mathcal{E}$} is comprised of a CNN or DINO backbone, one or multiple cascaded SalViTs, and a  SEM which is shared by all SalViT blocks, where the backbone is used for extracting raw tokens $\mathbf{F}^{\text{raw}}$, whereas the SalViT blocks are used for learning relations among these raw tokens, with the focus on foreground patches under the guidance of saliency. The output of each SalViT block is passed through a layer normalization (LN) and a fully-connected layer (FC), and then all outputs are concatenated along the channel mode to form $\mathbf{F}^{\text{vit}} = [\mathbf{F}^{1}; \cdots; \mathbf{F}^T]$, where $\mathbf{F}^t \in \mathbb{R}^{l \times l \times (d_{\text{vit}}/T)}$ and $T$ is the number of SalViT blocks. Finally, $\mathbf{F}^{\text{raw}}$ and $\mathbf{F}^{\text{vit}}$ are combined to form the output $\mathcal{E}(\mathbf{I})$.

%\vspace{0.1cm}
\noindent\textbf{Descriptor Net}, denoted by $\mathcal{D}$, contains dedicated $1\times 1$ conv. layer to adjust feature channel and a series of conv. layers with $(k=3, s=2, p=1)$ to reduce the resolution of the attentive feature maps down to desired sizes. Finally,  the output feature map is flattened into a feature vector that serves as a descriptor. %The intermediate layers contain a series of $3 \times 3$ conv. blocks which continuously reduce the feature map size. The output feature will be stretched into feature vector as descriptor.

\section{Further Details for Experiment Setup}
\subsection{Implementation Details}
\noindent\textbf{Our Approach:}
The M-SA of our SalViT has $6$ attention heads with each head using the linear projection to $64$ dimensional space. The localization net $\mathcal{G}$ uses three scales $S=\{8, 12, 16\}$. We use Adam as optimizer and set the learning rate to $1e\!-\!4$. Moreover, we set $W=2$ and $\eta=20$ by default in transductive FSKD.

\vspace{0.1cm}
\noindent\textbf{ProtoNet:} Original ProtoNet \cite{snell2017prototypical} forms the prototype for each class and classifies the query sample based on the nearest neighbor classifier (using the Euclidean distance). Similarly, we adapt ProtoNet to form support keypoint prototype (SKP) and compute the similarity between SKP and the feature vector at each position in query image. The location with the highest similarity is the detected keypoint.

\vspace{0.1cm}
\noindent\textbf{RelationNet:} We expand each SKP ($d$-dimensional vector) to be the same size as the query feature map ($l\times l\times d$), and then concatenate the expanded SKP and query feature map to form a pair of features for \emph{pixel-wise} relation learning using a relation module \cite{sung2018learning}. This approach is similar to FSS \cite{li2020fss}.

\vspace{0.1cm}
\noindent\textbf{WG:} Similar to \cite{gidaris2018dynamic,ge2021metacloth}, we predict the weights of keypoint localization net by using the feature averaging based mechanism and the attention-based mechanism. The attention-based mechanism composes the parameters of novel keypoint detector by %linearly
retrieving the parameters of base keypoint detector using attention (please see \cite{gidaris2018dynamic,ge2021metacloth} for more details). Thus, we can construct two approaches WG and WG (w/o Att.) for comparisons.

\vspace{0.1cm}
\noindent\textbf{Experiment Protocol:}
All-way $K$-shot ($K=1, 5$) protocol is used on all tasks. Specifically, each episode has $K$ support images. All base (or novel) keypoints serve as support keypoints. The episode is formed from same-species images. The model is trained on seen species with base keypoints while tested on unseen species with novel keypoints. We also evaluate performance for base keypoints (where available) on unseen species using identically-trained model.

% \noindent\textbf{Compared Methods:}
% \begin{itemize}
%     \item \emph{ProbIntr}
%     \item \emph{ProtoNet}
%     \item \emph{RelationNet}
%     \item \emph{WG}
%     \item \emph{TFA}
% \end{itemize}

\subsection{Keypoint Splits}\label{app:keypoint-splits}
\noindent
1) \textbf{Animal pose dataset.} \emph{Base keypoint set:} two ears, nose, four legs, four paws; \emph{Novel keypoint set:} two eyes, four knees. 2) \textbf{AwA pose dataset.} \emph{Base keypoint set:} two earbases, nose, neck base, neck end, throat base, throat end, tail, four legs, four paws, belly bottom, belly middle right, belly middle left; \emph{Novel keypoint set:} two eyes, four knees. 3) \textbf{CUB.} \emph{Base keypoint set:} beak, belly, back, breast, crown, two legs, nape, throat, tail; \emph{Novel keypoint set:} forehead, two eyes, two wings. 4) \textbf{NABird.} \emph{Base keypoint set:} beak, belly, back, breast, crown, nape, tail; \emph{Novel keypoint set:} two eyes, two wings. 5) \textbf{DeepFashion2.} \emph{Base keypoint set:} 158 total keypoints on 6 upper-body clothing categories; \emph{Novel keypoint set:} 136 total keypoints on 7 lower-body clothing categories\footnote{Please check \url{https://github.com/switchablenorms/DeepFashion2} for more details.}.

\comment{
\subsection{Auxiliary Keypoint Interpolation}\label{app:auxiliary-keypoint-interpolation}
The auxiliary keypoints are generated via interpolation $\mathcal{T}(t; [\mathbf{u}_1, \mathbf{u}_2])$ on a path whose end points are $[\mathbf{u}_1, \mathbf{u}_2]$, and the interpolation node is $t\in (0, 1)$. The interpolation path can be constructed by randomly sampling a number of paths formed by arbitrary body parts, or directly using default limbs. These two types of interpolation paths result in random version and default version of auxiliary keypoints. Following \cite{Lu_2022_CVPR}, we set $t={0.25, 0.50, 0.75}$ and the number of interpolation path to 6.
}

\section{Additional Ablation Studies}
\begin{table}[!tb]
  \centering
  \caption{Study on the number of selected pseudo labels $\eta$ in transductive FSKD. The harmonic scores of 1-shot keypoint detection are reported. ``Q'' with the number indicates the number of test query images.
  }
  \label{tab:tfskd-PL-number}
  \begin{tabular}{l|ll}
    %\hline
    Setting                       &Animal        & AwA  \\ \specialrule{.1em}{0em}{0em}
    \emph{Inductive}              &49.26         &68.18 \\
    \emph{Trans.} ($\eta=06$, Q15)&50.08 (+0.82) &68.89 (+0.71)\\
    \emph{Trans.} ($\eta=10$, Q30)&50.73 (+1.47) &69.13 (+0.95)\\
    \emph{Trans.} ($\eta=15$, Q45)&51.39 (+2.13) &69.84 (+1.66)\\
    \emph{Trans.} ($\eta=20$, Q60)&\textbf{52.49 (+3.23)} &\textbf{70.79 (+2.61)}\\
    %\hline
  \end{tabular}
\end{table}

  \noindent\textbf{Number of Pseudo Labels:} % we study this effects in transductive FSKD
  Table~\ref{tab:tfskd-PL-number} shows that increasing the number of selected pseudo labels yields better performance. Compared to the inductive inference, our model achieves improvements of 3.23\% and 2.61\% if selecting 20 pseudo labels out of 60 query images in Animal and AwA pose datasets, respectively. We also notice that increasing  the number of pseudo labels will eventually make the performance  saturated.

  \begin{figure*}[!tb]
    \centering
    \includegraphics[width=.95\linewidth]{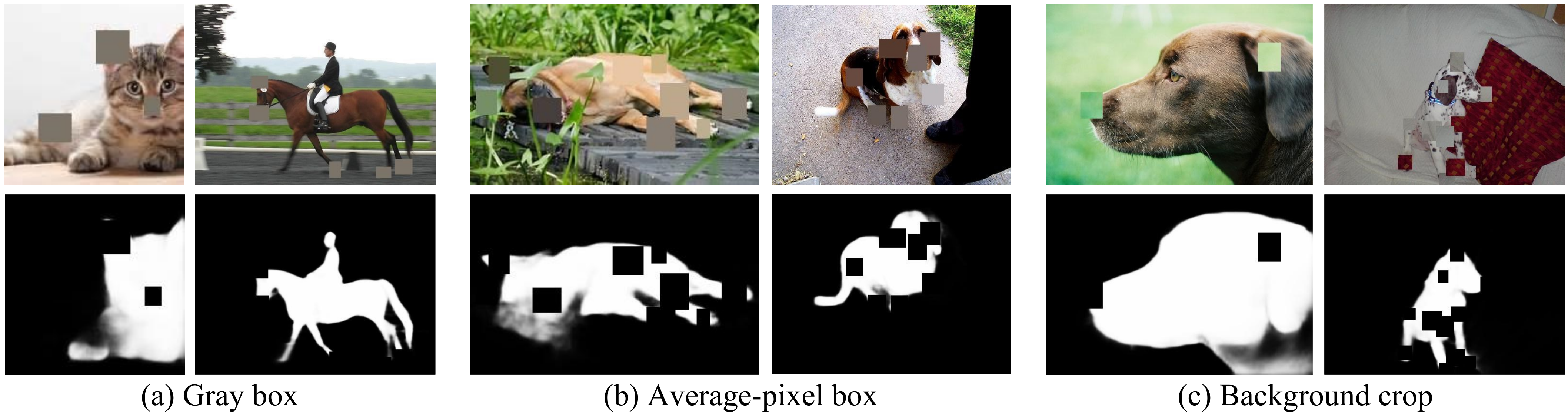}
    \caption{Visualization of three types of occlusions used in testing.
    }
    \label{fig:occlusions-types}
  \end{figure*}

  \begin{table}[!tb]
    \centering
    % \begin{minipage}{\linewidth}
    \caption{Testing results under occlusion of \textbf{gray box}. The model is trained using masks of \emph{gray box} with various masking strategies. The average PCK scores of novel keypoint detection on \emph{five} subproblems in Animal pose dataset are reported.
    }
    \label{tab:occlusion-graybox}
    % \small
    % \resizebox{\linewidth}{15mm}{
    \begin{tabular}{ccc|cccccc}
      % \toprule[1pt]
      % \hline
      RGB       & Sal.     & Feat.    &0\%  &30\% &50\% &100\% \\ \specialrule{.1em}{0em}{0em}%\hline
                &          &          &45.40&37.33&32.46&20.08 \\ %\hline
      \Checkmark&          &          &45.66&42.08&40.53&36.42 \\
      \Checkmark&\Checkmark&          &46.16&42.64&40.68&35.94 \\
      \Checkmark&\Checkmark&\Checkmark&45.81&42.06&40.16&34.53 \\
                % &\Checkmark&          &45.73&37.50&32.22&19.56 \\
                &\Checkmark&\Checkmark&45.86&37.69&32.80&20.14 \\
      % \bottomrule[1pt]
      % \hline
    \end{tabular}
    % }  % resizebox
  \end{table}
  \begin{table}[!tb]
  % \vspace{0.1cm}
  % \end{minipage}
  % \begin{minipage}{\linewidth}
    \centering
    \caption{Testing results under occlusion of \textbf{average-pixel box}. The model is trained using masks of \emph{gray box} with various masking strategies. The average PCK scores of novel keypoint detection on \emph{five} subproblems in Animal pose dataset are reported.
    }
    \label{tab:occlusion-avgpixelbox}
    % \small
    % \resizebox{\linewidth}{15mm}{
    \begin{tabular}{ccc|cccccc}
      % \toprule[1pt]
      % \hline
      RGB       & Sal.     & Feat.    &0\%  &30\% &50\% &100\% \\ \specialrule{.1em}{0em}{0em}%\hline
                &          &          &45.16&38.34&34.06&22.60 \\ %\hline
      \Checkmark&          &          &45.54&42.21&40.80&36.40 \\
      \Checkmark&\Checkmark&          &45.90&42.58&40.94&36.33 \\
      \Checkmark&\Checkmark&\Checkmark&45.94&42.40&40.41&35.35 \\
                % &\Checkmark&          &45.55&38.46&34.14&22.20 \\
                &\Checkmark&\Checkmark&46.38&38.71&34.39&22.81 \\
      % \bottomrule[1pt]
      % \hline
    \end{tabular}
    % } % resizebox
  \end{table}
  \begin{table}[!tb]
  % \vspace{0.1cm}
  % \end{minipage}
  % \begin{minipage}{\linewidth}
    \centering
    \caption{Ablations of various masking and alignment (MAA) strategies under testing occlusion of \textbf{gray box}. The results are the average over five subproblems on the Animal pose dataset.  Non-Occl means adding the prediction loss from non-occlude images. Y: applied; N: not applied.
    }
    \label{tab:MAA-graybox}
    \begin{tabular}{cl|ccc}
      %\hline % average
      Mask & Alignment        &0\%  &50\% &100\% \\ \specialrule{.1em}{0em}{0em}%\hline
        N  & N                &45.40&32.46&20.08 \\ %\hline
        Y  & N                &46.16&40.68&35.94 \\ %\hline
        % Y  & MAE              &     &     &      \\
        Y  & SimMIM \cite{xie2022simmim}&45.87&40.59&36.78 \\
        Y  & Non-Occl Loss          &46.20&40.16&36.14 \\
        Y  & Feat. Align (MMD)      &45.76&40.05&35.50 \\
        Y  & Feat. Align ($\ell_2$) &46.40&41.25&36.75 \\
        Y  & Feat. Align ($\ell_1$) &\textbf{46.59}&40.31&36.00 \\
        Y  & Prob. Align            &46.44&\textbf{41.52}&\textbf{38.40} \\
        % Y  & Feat. (MSE)+Prob.      &45.78&40.51&36.95 \\
      %\hline
    \end{tabular}
  \end{table}
  \begin{table}[!tb]
  % \vspace{0.1cm}
  % \end{minipage}
  % \begin{minipage}{\linewidth}
    \centering
    \caption{Ablations of various masking and alignment (MAA) strategies under testing occlusion of \textbf{average-pixel box}. The results are the average over five subproblems on the Animal pose dataset. Non-Occl means adding the prediction loss from non-occlude images. Y: applied; N: not applied.
    }
    \label{tab:MAA-avgpixelbox}
    \begin{tabular}{cl|ccc}
      %\hline % average
      Mask & Alignment        &0\%  &50\% &100\% \\ \specialrule{.1em}{0em}{0em}%\hline
        N  & N                &45.16&34.06&22.60 \\ %\hline
        Y  & N                &45.90&40.94&36.33 \\ %\hline
        % Y  & MAE              &     &     &      \\
        Y  & SimMIM \cite{xie2022simmim}&45.82&41.21&37.02 \\
        Y  & Non-Occl Loss          &45.88&40.84&36.13 \\
        Y  & Feat. Align (MMD)      &45.24&40.16&35.70 \\
        Y  & Feat. Align ($\ell_2$) &46.01&40.77&37.14 \\
        Y  & Feat. Align ($\ell_1$) &46.20&40.93&36.25 \\
        Y  & Prob. Align            &\textbf{46.29}&\textbf{42.24}&\textbf{38.54} \\
        % Y  & Feat. (MSE)+Prob.      &45.50&41.23&36.68 \\
      %\hline
    \end{tabular}
  % \end{minipage}
  \end{table}

  \begin{figure*}[!tb]
    \centering
    \includegraphics[width=\linewidth]{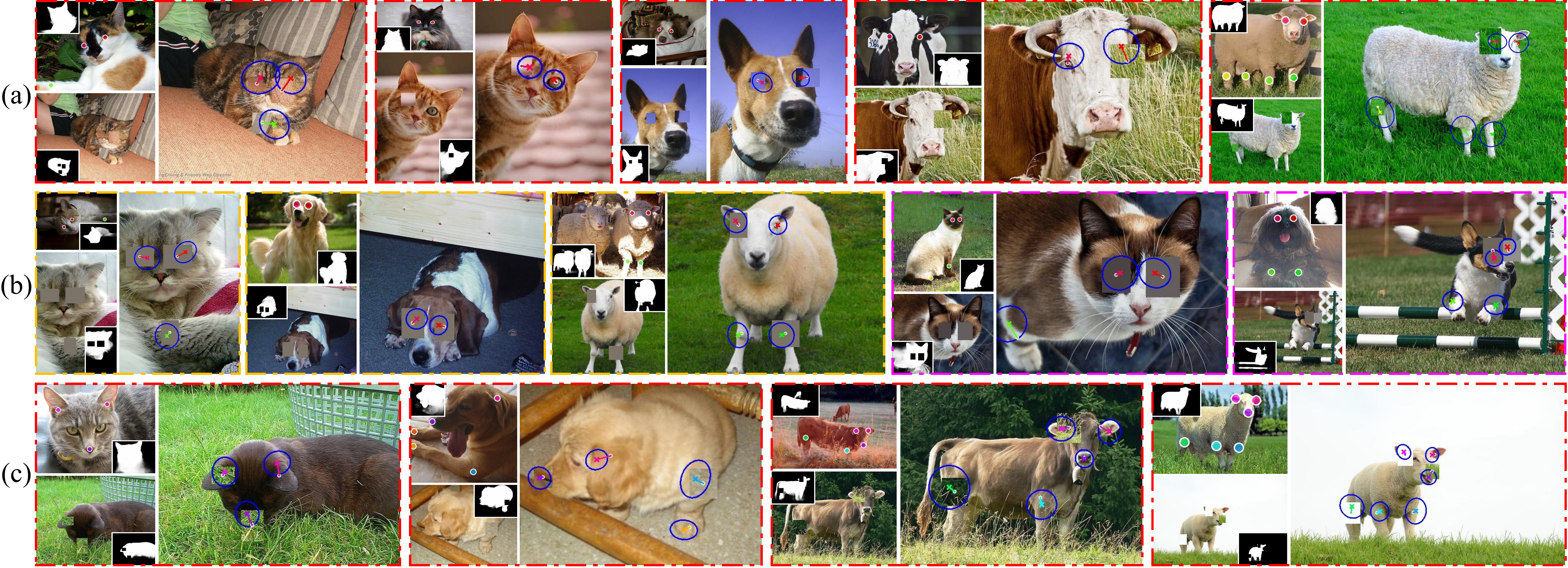}
    \caption{Examples of few-shot keypoint detection for unseen species under occlusions. Rows (a) and (b) are for detecting novel keypoints while (c) is for detecting base keypoints. The dashed boxes in red, yellow, and pink represents using occlusion types of \emph{background crop}, \emph{gray box}, and \emph{average-pixel box}, respectively. Within each dashed box, the left-top image is the support, the left-bottom one is the query, and the right shows the detected keypoints with uncertainty.
  }
    \label{fig:fskd-vis-under-occlusion}
  \end{figure*}

  \section{Further Results on FSKD Under Occlusion}\label{app:more-results-of-occlusions}
  \noindent\textbf{Occlusion Visualization:}
  The occlusion types regarding \emph{gray box}, \emph{average-pixel box}, and \emph{background crop} are visualized in Fig.~\ref{fig:occlusions-types}, which shows that the \emph{background crop} (a crop extracted from the background and transplanted into the foreground) is the best type of simulated occlusion. For example, the ``nose'' of the dog in Fig.~\ref{fig:occlusions-types}(c) is realistically occluded. % in high fidelity.

  \vspace{0.1cm}
  \noindent\textbf{Testing with Various Occlusions:}
  The testing results under occlusion of \emph{gray box} and occlusion of \emph{average-pixel box} are shown in Tables~\ref{tab:occlusion-graybox} and \ref{tab:occlusion-avgpixelbox}, respectively. Compared to Table~\ref{tab:occlusion-bgbox} which uses occlusions of \emph{background crop} type in the main paper, this experiment shows that the scores are higher if one uses  \emph{gray box} or \emph{average-pixel box} as testing occlusions (\eg, when both RGB and saliency maps are masked, comparing 35.94\% in Table~\ref{tab:occlusion-graybox} and 36.33\% in Table~\ref{tab:occlusion-avgpixelbox} to 29.77\% in Table~\ref{tab:occlusion-bgbox}), because \emph{gray box} and \emph{average-pixel box} are simpler and induce lesser domain shift compared with the training masks of \emph{gray box}. Regardless of occlusion types, the FSKD with SalViT model equipped with our masking strategy outperforms regularly-trained model (\ie, no masking applied) by a large margin, which validates that the masking strategy is useful and the image-masked model has learned to handle occlusions to some degree.

  \vspace{0.1cm}
  \noindent\textbf{MAA Strategy:}
  % Test under graybox, average-pixel box. We need to compare with SIMMIM.
  Firstly, we provide two baselines: i) the normally-trained model without masking input (\ie, RGB + saliency) and without alignment; ii) the model trained with masking input but without the alignment strategy. Then, we investigate different alignment approaches to improve handling occlusions. SimMIM~\cite{xie2022simmim}  recovers the raw pixel values of randomly masked image patches with a linear layer under the $\ell_1$ loss, which can be regarded as alignment in the image space. Thus, we select SimMIM for comparisons. Since we have access to both occluded and non-occluded views of input, the prediction loss from non-occluded images can be added into the joint optimization, in addition to $\mathcal{L}_{\text{ms}}^{\text{occ}}$ in Eq.~\ref{eq:total-loss} (in the main paper). The ``Non-Occl Loss'' and ``Occl Loss'' are forced to make the same predictions, thus one can think they implicitly perform alignment between two data views. Finally, we compare variants of feature alignment based on various distance metrics ($\ell_1$, $\ell_2$, and and MMD \cite{borgwardt2006integrating}), and probability alignment. From Tables~\ref{tab:MAA-graybox} and \ref{tab:MAA-avgpixelbox}, we observe: 1) Choosing which modality/signal to align is crucial. The probability alignment is very effective and outperforms other approaches; 2) Using SimMIM (alignment in the image space) and feature alignment are useful under severe occlusions of 100\%. Feature alignment with the $\ell_1$ or $\ell_2$ metric is better than using the MMD metric; 3) The testing scores under occlusions of \emph{gray box} (Table~\ref{tab:MAA-graybox}) and \emph{average-pixel box} (Table~\ref{tab:MAA-avgpixelbox}), again, are higher than under occlusions of \emph{background crop} (Table~\ref{tab:MAA-bgcrop}, in main paper). These results manifest that the simple MAA can improve  occlusion handling in our FSKD with SalViT model.

  \noindent\textbf{Visualization Results:} % visualization of FSKD under occlusion
  Fig.~\ref{fig:fskd-vis-under-occlusion} shows the visualization of few-shot keypoint detection under three types of occlusions. The query keypoint is randomly occluded with the probability of 0.3. As one can see, our model with simple MAA can effectively infer the locations of novel and base keypoints for unseen species, even though the body part is entirely occluded with visually realistic occlusions. Moreover, the estimated uncertainty (\ie, blue ellipses in Fig.~\ref{fig:fskd-vis-under-occlusion}) still correlates well with the shape of body parts under occlusion setting.

  % \section{Few-shot Fine-grained Classification}
  % same setup to FGVR downstream task in our CVPR'22 paper

  \begin{figure}[!tb]
    \centering
    \includegraphics[width=\linewidth]{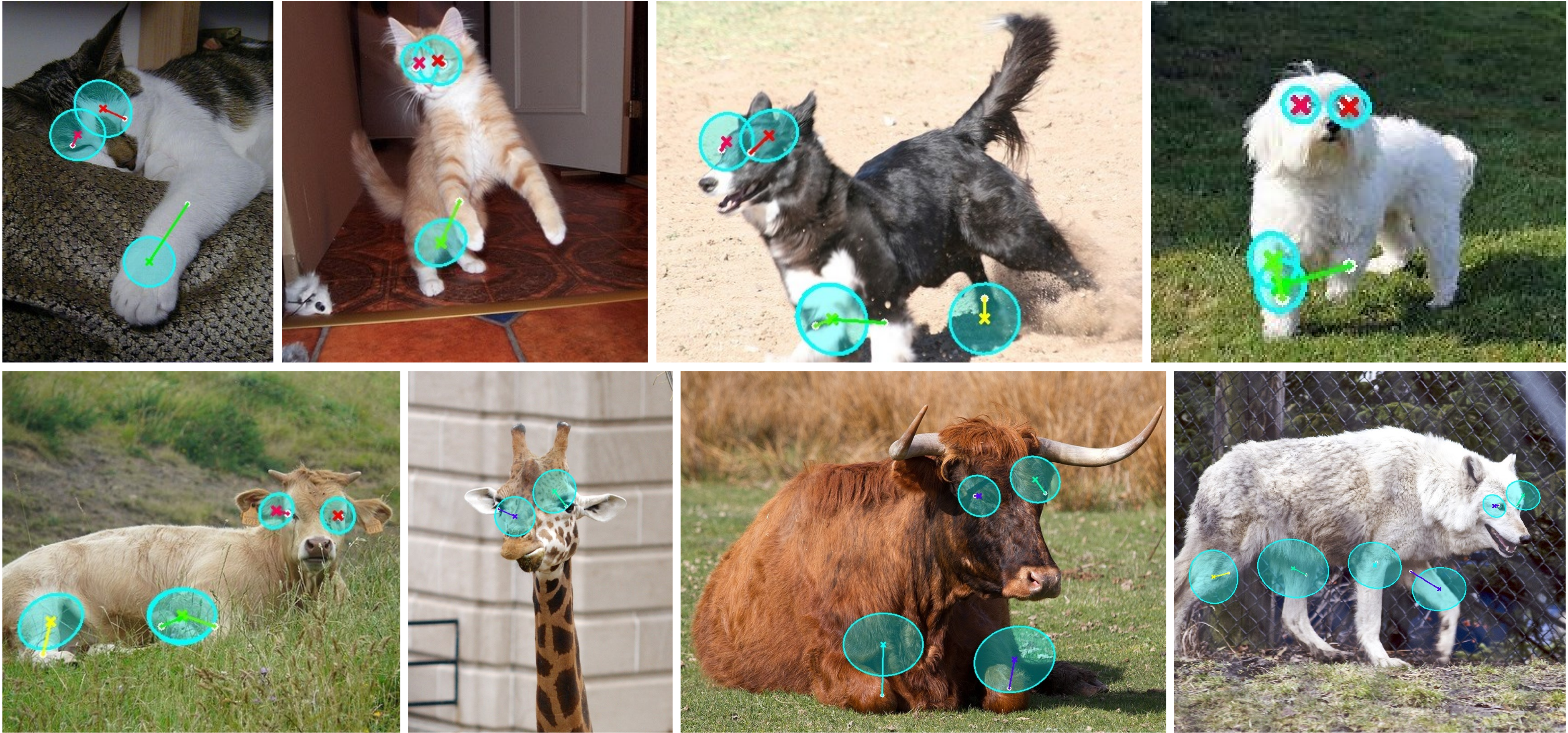}
    \caption{Selected examples of failure cases for novel keypoint detection on unseen species. The dots refer to GT query keypoints. The tilted cross centered at an ellipse reflects the keypoint prediction with uncertainty. The line that connects GT and predicted keypoint reflects the localization error.
    }
    \label{fig:failure-cases}
    % \vspace{-7pt}
  \end{figure}

  \section{Failure Cases and Future Work}
  % some limitations such as relying on external saliency detector
  % show some failure cases
  Development of a keypoint learner that can efficiently detect arbitrary novel keypoints given a few exemplars remains an open question. Although our SalViT partially addresses FSKD by providing the informative relations within foreground patches for keypoint inference, FSKD can fail when  colors and textures are indistinguishable. For example, the body part ``knee" of the cat has been wrongly detected as ``paw'', as shown in the first image of Fig.~\ref{fig:failure-cases}. Moreover, as FSKD is a very hard problem (arguably harder than standard FSL and FSOD), the keypoint predictions are usually imperfect (see localization error in Fig.~\ref{fig:failure-cases}). Although our modelled uncertainty mitigates this issue well, we hope to further improve the localization accuracy in our future works.

\end{document}